\documentclass{article}

\usepackage{microtype}
\usepackage{graphicx}
\usepackage{subfigure}
\usepackage{booktabs} % for professional tables
\usepackage{makecell}
\usepackage[table,xcdraw]{xcolor}
\usepackage{exscale}

\usepackage{amsmath,amsfonts,bm}

\def\Acal{{\mathcal{A}}}

\def\Dcal{{\mathcal{D}}}

\def\Pcal{{\mathcal{P}}}

\def\Scal{{\mathcal{S}}}

\def\eqref#1{equation~\ref{#1}}
\def\1{\bm{1}}

\DeclareMathAlphabet{\mathsfit}{\encodingdefault}{\sfdefault}{m}{sl}
\SetMathAlphabet{\mathsfit}{bold}{\encodingdefault}{\sfdefault}{bx}{n}

\usepackage{multirow}

\usepackage{hyperref}
\usepackage[accepted]{icml2024}

\usepackage{amsmath}
\usepackage{amssymb}
\usepackage{mathtools}
\usepackage{amsthm}
\usepackage{hhline}
\usepackage[capitalize,noabbrev]{cleveref}

\theoremstyle{plain}
\newtheorem{theorem}{Theorem}[section]

\theoremstyle{definition}
\newtheorem{definition}[theorem]{Definition}

\theoremstyle{remark}

\usepackage[textsize=tiny]{todonotes}
\usepackage{subcaption}

\icmltitlerunning{Temporal Logic Specification-Conditioned Decision Transformer for Offline Safe Reinforcement Learning}

\begin{document}

\twocolumn[
% \icmltitle{Temporal Logic Specification-Conditioned Decision Transformer}
\icmltitle{Temporal Logic Specification-Conditioned Decision Transformer\\ for Offline Safe Reinforcement Learning}



\icmlsetsymbol{equal}{*}

\begin{icmlauthorlist}
\icmlauthor{Zijian Guo}{se}
\icmlauthor{Weichao Zhou}{ece}
\icmlauthor{Wenchao Li}{ece}

\end{icmlauthorlist}

\icmlaffiliation{se}{Division of Systems Engineering, Boston University}
\icmlaffiliation{ece}{Department of Electrical and Computer Engineering, Boston University}

\icmlcorrespondingauthor{Zijian Guo}{zjguo@bu.edu}
% \maketitle
% \authors


\vskip 0.3in
]



\printAffiliationsAndNotice{}  % leave blank if no need to mention equal contribution

\begin{abstract}
Offline safe reinforcement learning (RL) aims to train a constraint satisfaction policy from a fixed dataset.
Current state-of-the-art approaches are based on supervised learning with a conditioned policy. 
However, these approaches fall short in real-world applications that involve complex tasks with rich temporal and logical structures.
In this paper, we propose temporal logic Specification-conditioned Decision Transformer (SDT), a novel framework that harnesses the expressive power of signal temporal logic (STL) to specify complex temporal rules that an agent should follow and the sequential modeling capability of Decision Transformer (DT).
Empirical evaluations on the \texttt{DSRL} benchmarks demonstrate the better capacity of SDT in learning safe and high-reward policies compared with existing approaches. In addition, SDT shows good alignment with respect to different desired degrees of satisfaction of the STL specification that it is conditioned on.
\end{abstract}

\section{Introduction}
\label{sec:into}

Offline safe reinforcement learning (RL) is the problem of learning a policy to achieve high rewards while keeping the cost of constraint violation below a specified threshold from a fixed dataset without further interactions with the RL environment.
It is preferable to online safe RL when data collection is expensive and inefficient~\cite{gu2022review, guo2023sample}.
% Transformers have made great advancements across various domains, such as natural language processing~\cite{tunstall2022natural} and computer vision~\cite{khan2022transformers}.
% The Decision transformers (DT)~\cite{chen2021decision} successfully adopts the transformers architecture~\cite{vaswani2017attention} to offline reinforcement learning (RL) and shows notable performance.
% As safety is the premier for many real-world applications and the advantage of not requiring interaction with environments, offline safe RL is gaining attention~\cite{ma2021conservative, xu2021safely, zheng2024safe} and 
Various methods have been proposed and shown promising performance in attaining high rewards and satisfying relatively simple constraints in applications such as autonomous driving~\cite{gu2022constrained, lin2023safety}, robotics~\cite{brunke2022safe}, and healthcare~\cite{kondrup2023towards, zhang2024balancing}.
However, real-world tasks often require agents to follow complex temporal and logical constraints.
For example, autonomous-driving vehicles must come to a complete halt at an intersection with a stop sign, pause briefly, and proceed through it only if no other cars are present.
It is challenging to define cost functions that appropriately capture such constraints.

% Signal temporal logic (STL) has the advantages of precise temporal specification, flexibility in defining safety criteria over time, and the ability to incorporate both spatial and temporal aspects into constraints~\cite{donze2013signal}.
% In cyber-physical systems, the \textit{specification} of STL is used for describing temporal properties such as \textit{safety} (never visit a bad state), \textit{liveness} (eventually visit a good state), \textit{sequentiality} (visit state A then state B), and their arbitrarily elaborate combinations~\cite{kress2009temporal, coogan2017formal, madsen2018metrics, vasile2017minimum}.
% Due to its expressiveness, STL is often used for controller verification and synthesis combined with control theories~\cite{sahin2020autonomous, kurtz2021more, zhang2023modularized, meng2023signal}.
Temporal logic (TL), 
on the other hand, 
provides a formalism for expressing the behaviors of systems over time, such as \textit{safety} (e.g. never visit a bad state), \textit{liveness} (e.g. eventually visit a good state), \textit{sequentiality} (e.g. visit state A before visiting state B), and their arbitrarily elaborate combinations~\cite{donze2013signal, tabuada2015robust, coogan2017formal, vasile2017minimum, madsen2018metrics}.
% \li{add citations for LTL}.
Signal temporal logic (STL) is a variant of TL that can be used to specify properties over dense-time real-valued signals such as trajectories~\cite{maler2004monitoring}. 
A unique characteristic of STL is that it admits a quantitative semantics -- a \textit{robustness value} that quantifies the degree to which a given trajectory satisfies an STL formula~\cite{fainekos2009robustness, donze2010robust}.
Due to this quantitative semantics and its expressiveness, STL has found a wide range of applications ranging from controller verification and synthesis~\cite{eddeland2017objective, sahin2020autonomous, kurtz2021more, dawson2022robust, zhang2023modularized} to reinforcement learning~\cite{balakrishnan2019structured, bozkurt2020control, zhang2021model}.

In this paper, we propose to leverage STL to solve offline safe RL problems that involve complex temporal constraints. 
To tackle the challenges of learning from a fixed dataset, we build on the Decision Transformer (DT) model~\cite{chen2021decision} which leverages the self-attention mechanism in Transformers~\cite{vaswani2017attention} to handle long-range dependencies and has demonstrated competitive performance in offline RL settings. 
%The transformer architecture implicitly forms the state-return/cost associations via the similarity of the query and key vectors within the model.
%Recent works have extended DT to allow conditioning on goals~\cite{emmons2022rvs}, costs~\cite{liu-icml23-cdt}, and with online finetuning~\cite{zheng2022online}.
A crucial insight of our work is that, 
while the notions of goals, returns, and costs are tied to RL, the sequence modeling and generation framework of DT is more general. 
Our novel framework, called 
temporal logic Specification-conditioned Decision Transformer (SDT), judiciously combines the sequential modeling capability of DT with the expressive power of STL
to learn high-performance and safe policies.
%In the offline RL setting, the Decision Transformers (DT) \cite{chen2021decision} successfully adopts the transformers architecture~\cite{vaswani2017attention} and demonstrates its capability to relate sequential information and produce accurate outputs.
%In this paper, we propose the temporal logic Specification-conditioned Decision Transformers (SDT) to achieve safety and high-reward performance by utilizing the sequential modeling of DT to learn from an offline dataset, with a specific focus on addressing the temporal dimension of constraints expressed in the robustness value of STL.
Our main contributions are summarized as follows.

\vspace{-2mm}
\begin{itemize}
    % \item We study the offline safe RL problem from the supervised learning perspective and propose to train decision transformers to satisfy temporal logic specification.
    \item We study the offline safe RL problem from the supervised learning perspective and propose SDT which enables conditioning on STL specifications in DT. 
    % As far as we are aware, 
    % To the best of our knowledge,
    % this is the first work that incorporates STL in the offline RL setting to accomplish temporal constraints.
    SDT is the first work that incorporates STL to satisfy temporal constraints in offline safe RL settings.
    \vspace{-1mm}
    \item We examine the capacity of autoregressive learning with the quantitative semantics of STL.
    Our method introduces two key input tokens: the prefix and suffix robustness values that leverage different portions of a trajectory to provide complementary information.
    \vspace{-1mm}
    \item Our comprehensive experiments show that i) SDT outperforms multiple baselines both in safety and task performance by a large margin;  
    ii) SDT can generalize to different robustness value thresholds and configurations without re-training the policy.
\end{itemize}

\vspace{-4mm}
\section{Related Work}
\label{sec:related-work}

\textbf{Constrained optimization for offline safe RL.}
Offline safe RL is the intersection of safe RL~\cite{achiam2017constrained, xu2021crpo} and offline RL~\cite{levine2020offline, prudencio2023survey}, where agents aim to balance safety and efficiency with reward, cost, and cost threshold information by learning from fixed trajectories.
% ~\cite{liu2023datasets}.
Most of the recent works formulate the safe RL problem as a constrained optimization problem.
The Distribution Correction Estimation (DICE) family of RL algorithms~\cite{kostrikov2021offline, lee2021optidice, lee2022coptidice} optimizes the policy based on explicitly estimating the distributional shift between the target policy and the offline data distribution.
The primal-dual method with function approximation optimizes iteratively between the policy and the lagrangian multiplier used to penalize constraint violations~\cite{le2019batch, chen2022near, xu2022constraints, polosky2022constrained, hong2023primal}.
However, the aspect of addressing temporal constraints remains underexplored.

\textbf{Conditioned RL.}
Reward-Conditioned Supervised Learning (RCSL) represents a burgeoning category of algorithms that learn action distribution based on future return statistics via supervised learning.
This idea is first proposed for the online RL setting~\cite {schmidhuber2019reinforcement, kumar2019reward, peng2019advantage}.
% \li{need better transition here}
Decision Transformer (DT) and its variants~\cite{chen2021decision, furuta2021generalized, zheng2022online, yamagata2023q, wang2023critic, hu2023graph} extend this idea to the offline RL setting by using return-to-go, i.e., cumulative future reward, as the conditional inputs and modeling trajectories with casual transformers~\cite{vaswani2017attention}.
% \li{need better transition here}
Instead of complex models, recent findings show that with careful policy tuning, simple multi-layered neural networks can match transformers' performance~\cite{emmons2021rvs, brandfonbrener2022does}. 
Several works extend reward-conditioned RL methods in the offline safe RL setting by additionally conditioning on cost-to-go, i.e., cumulative future cost, and achieve not only safety and performance but also generalization to different cost thresholds~\cite{liu2023constrained, zhang2023saformer}.
Our work adopts formal specifications in the conditions to characterize temporal properties that are difficult to be captured by standard reward or cost functions.

\textbf{STL as reward (or cost) functions in RL.}
A well-defined reward (or cost) function is essential for RL agents to accomplish the underlying tasks~\cite{amodei2016concrete,zhou2022aaai}.
Due to its rich syntax, STL can be used to specify high-level goals and safety requirements in RL tasks and also evaluate the behaviors of agents in conforming to the specifications combined with model-free methods~\cite{li2017reinforcement, toro2018teaching, camacho2019ltl, balakrishnan2019structured}. 
Several model-based methods have been proposed~\cite{kapoor2020model, cohen2021model} to improve sample efficiency and promote safe training, but the inherent trial-and-error process of RL still poses a significant risk of performing unsafe actions.
This work focuses on the offline setting to avoid unsafe explorations during training.

\vspace{-2mm}
\section{Preliminaries}
\label{sec:preliminaries}

\subsection{Offline Safe RL}
We consider learning in a finite horizon Markov decision process (MDP) which can be described by the tuple $(\Scal, \Acal, \Pcal, r)$, 
where $\Scal$ is the state space, $\Acal$ is the action space, $\Pcal$ is the transition function and $r$ is the reward function.
A trajectory comprises a sequence of states and actions $\tau = \{s_t, a_t\}_{t=1}^T$ with length $|\tau| = T$.
The goal of RL is to learn a policy $\pi$ that maximizes the expected cumulative reward $\mathbb{E}_{\tau \sim \pi}  \big[\sum_{t=1}^T r(s_t, a_t)]$.

A Constrained MDP (CMDP)~\cite{altman2021constrained} augments MDP with a cost function $c$, and is commonly used to model safe RL whose goal is to maximize the cumulative reward while limiting the cumulative cost to a threshold $d$:
\begin{equation}
    \small
   \max_{\pi} \mathbb{E}_{\tau \sim \pi}  \big[\sum_{t=1}^T r(s_t, a_t), \text{ s.t. } \mathbb{E}_{\tau \sim \pi} \big[\sum_{t=1}^T c(s_t, a_t)] \leq d. 
   \label{eq:safe-rl}
\end{equation} 
In the offline setting, the agent cannot interact with the environment and can only access a fixed dataset $\Dcal = \{\tau_i\}_{i=1}^n$ consisting of trajectories collected by unknown policies.

% where $\Scal$ is the state space, $\Acal$ is the action space, $\Pcal:\Scal \times \Acal \times \Scal \xrightarrow{} [0, 1]$ 
% is the transition function, $r:\Scal \times \Acal \times \Scal \xrightarrow{} \mathbb{R}$ is the reward function, $c:\Scal \times \Acal \times \Scal \xrightarrow{} \mathbb{R}$ is the cost function that characterizes the cost for violating the constraint, and $\mu_0: \Scal \xrightarrow[]{} [0,1]$ is the initial state distribution.
% The cumulative reward and cost of a trajectory $\tau = \{s_1, a_1, ..., s_T, a_T\}$ with length $|\tau| = T$ is calculated as $R(\tau) = \sum_{t=1}^T r(s_t, a_t)$ and $C(\tau) = \sum_{t=1}^T c(s_t, a_t)$.
% The goal of safe RL is to find the policy $\pi: \Scal \times \Acal\rightarrow [0,1]$ that maximizes the cumulative reward while limiting the cumulative cost to a specified threshold $d \in \mathbb{R}$:
% % \vspace{-2mm}
% \begin{equation}
%    \max_{\pi} \mathbb{E}_{\tau \sim \pi}  \big[R(\tau)], \quad s.t. \quad \mathbb{E}_{\tau \sim \pi} \big[C(\tau)] \leq d. 
%    \label{eq:safe-rl}
% \end{equation} 
% % \vspace{-2mm}
% In the offline setting, the agent cannot interact with environments but only access a fixed dataset $\Dcal = \{\tau_i\}_{i=1}^n$ consisting of trajectories collected by unknown policies.

\vspace{-2mm}
\subsection{Decision Transformers}
The Decision Transformer (DT) model~\cite{chen2021decision}
tackles offline RL as a sequential modeling problem.
Unlike the majority of prior RL approaches that estimate value functions and parameterize a single state-conditioned policy $\pi(a|s)$, DT outputs predicted actions from a sequence of return-to-go $\mathbf{R}_{t-K:t}=$ $\{R_{t-K}$, $...$, $R_t\}$, where $R_t = \sum_{t'=t}^T r_t(s_{t'}, a_{t'})$ is the cumulative reward from time-step $t$ and $K$ is the context length; states $\mathbf{s}_{t-K:t}=$ $\{s_{t-K}$, $...$, $s_t\}$; and actions $\mathbf{a}_{t-K:t}=\{a_{t-K}, ..., a_t\}$.
A DT's policy is parametrized by the GPT architecture \cite{radford2018improving} with a causal self-attention mask.
It generates a deterministic action $\pi_{DT}(R_{t-K:t}, s_{t-K:t}, a_{t-K:t})$ at each time-step $t$.
The policy is trained by minimizing the loss between the predicted and ground-truth actions. 

\vspace{-2mm}
\subsection{STL Specification}
% STL and robustness value semantics
% Signal temporal logic (STL)~\cite{donze2010robust} is a real-time logic, typically interpreted over signals (i.e., states in this work) over a dense-time domain that takes values in a continuous metric space. 
% The syntax of our fragment of STL is formally defined as follows:
Signal Temporal Logic (STL) is a formal logic used to specify and monitor temporal properties of continuous signals~\cite{donze2010robust}.
In this paper, we consider states at discrete time-steps as signals under discrete-time sampling.
The syntax of STL is as follows.
%The desired behavior of agents can be described by using the STL syntax:
% In this paper, the desired behavior of agents is described by a signal temporal logic (STL) fragment with the following syntax:
% \vspace{-4mm}
\begin{equation}
% \vspace{-2mm}
    \centering
    \small
    \begin{aligned}
        \phi := \ \top \ | \ & \mu_c \ | \ \neg \phi \ | \ \phi \land \psi \ | \ \phi \lor \psi \ | 
        % & \mathbf{G}_{[a, b]}\phi \ | \ \mathbf{F}_{[a, b]}\phi \ | \ \phi \mathbf{U}_{[a, b]} \psi \ | \ \phi \Rightarrow \psi
        \ \phi \Rightarrow \psi \ | \ \\
        &
        \mathbf{G}_{[t_1, t_2]}\phi \ | \ \mathbf{F}_{[t_1, t_2]}\phi \ | \
        \phi \mathbf{U}_{[t_1, t_2]} \psi
    \end{aligned}
    \label{eq:stl-def}
\end{equation}
% \vspace{-4mm}
where $\mu_c$ is a predicate of the form $\mu(s) < c$ where $\mu(\cdot): \Scal \xrightarrow{} \mathbb{R}$ and $c \in \mathbb{R}$ is a constant; $\phi$ and $\psi$ are STL formulas; $t_1, t_2 \in \mathbb{Z}^{+}$ denote two sequential time steps.
$\top$ represents True.
%The Boolean operators $\neg$, $\land$, and $\lor$ are negation, conjunction (i.e., \textit{and}) and disjunction (i.e., \textit{or}) respectively.
%The implication operator $\Rightarrow$ in $\phi\Rightarrow \psi$ is equivalent to $\phi \Rightarrow \psi \equiv \neg \phi \lor \psi$.
The temporal operators $\mathbf{G}$, $\mathbf{F}$, and $\mathbf{U}$ refer to \textit{Globally} (i.e., always), \textit{Finally} (i.e., eventually), and \textit{Until}.
The \textit{quantitative semantics}~\cite{donze2010robust} of STL as shown in Eq.~(\ref{eq:stl-rob}) quantifies the degree to which a trajectory $\tau$ satisfies or violates an STL formula $\phi$ at each time step $t$ through the robustness value, denoted as $\rho(\tau, t, \phi)$.
% For simplicity, we will use $\mathbf{s}$.
% Formally, the robustness value is defined using the following recursive semantics:
% \vspace{-1mm}
% \vspace{-4mm}
\begin{equation}
    \centering
    \small
    \begin{aligned}
         \rho(\tau, t, \top) & = \rho_{max} \quad \text{where } \rho_{max} > 0 \\
         \rho(\tau, t, \mu_c) & = c - \mu(s_t) \\
         \rho(\tau, t, \neg\phi) & = -  \rho(\tau, t, \phi) \\
         \rho(\tau, t, \phi_1 \land \phi_2) & = \min\bigl( \rho(\tau, t, \phi_1),  \rho(\tau, t, \phi_2)\bigl) \\
         \rho(\tau, t, \phi_1 \lor \phi_2) & = \max\bigl( \rho(\tau, t, \phi_1),  \rho(\tau, t, \phi_2)\bigl) \\
         \rho(\tau, t, \phi \Rightarrow \psi) & = \max\bigl(- \rho(\tau, t, \phi),  \rho(\tau, t, \psi)\bigl) \\
         \rho(\tau, t, \mathbf{G}_{[t_1, t_2]}\phi) &= \min_{t' \in [t+t_1, t+t_2]} \bigl(\rho(
         \tau, t', \phi)\bigl) \\
         \rho(\tau, t, \mathbf{F}_{[t_1, t_2]}\phi) &= \max_{t' \in [t+t_1, t+t_2]} \bigl(\rho(\tau, {t'}, \phi)\bigl) \\
         \rho(\tau, t, \phi \mathbf{U}_{[t_1, t_2]}\psi) & = \max_{t' \in [t+t_1, t+t_2]} \min \bigl(\rho(\tau, {t'}, \psi),  \\[-4pt]
        & \quad \quad \quad \min_{t'' \in [t, t']} \rho(\tau, {t''}, \phi)\bigl)
    \end{aligned}
    \label{eq:stl-rob}
    % \vspace{-1mm}
\end{equation}
% \vspace{-4mm}
% where $t + t_1 < t + t_2 \leq T$. 
%The robustness value measures the degree of a trajectory satisfying or violating an STL specification. 
%A positive robustness value indicates satisfaction of the formula, and a higher robustness value indicates stronger satisfaction (and vice versa).
The sign of the robustness value indicates whether the specification is satisfied or not, while a higher value indicates stronger satisfaction and vice versa.
A specification can contain multiple predicates whose values can be arbitrarily scaled, yet the value of the overall specification maintains strict semantics.
% The robustness value of the predicate measures how much smaller $\mu(s_t)$ is than $c$. 
% For the \textit{and} operation,
% % denoted as $\phi \land \psi$, 
% the requirement is for both $\phi$ and $\psi$ to be true; thus, the robustness value takes the minimum of the two robustness values. 
% In contrast, the \textit{or} operation seeks the truth of either $\phi$ or $\psi$, so the maximum of the two robustness values is used in this case. 
% When dealing with the \textit{Globally} operation, the minimum of the robustness values across all states is considered, indicating that each state should meet the specification.
% Conversely, the \textit{Finally} operation, which requires at least one state to satisfy the specification, uses the maximum robustness value.

% \subsection{Offline Safe RL via Supervised Learning}

% When dealing with the \textit{Until} operation, the term $min_{t''\in[t,t']} \rho(\mathbf{s}_{t''}, \phi)$ is associated with the necessity of $\phi$ being true for all times within the interval $[t, t']$, where $t'$ itself lies within the interval $[t + t_1, t + t_2]$. 
% The subsequent min operation aligns with the requirement for $\psi$ to be true at time $t'$.
% The max operation that follows ensures that the conditions for both $\phi$ and $\psi$ are met at least once during the interval $[t + t_1, t + t_2]$.

% \vspace{-2mm}
\section{Method}
\label{sec:method}

In this section, we first outline the formulation to solve the offline safe RL problem through supervised learning. 
Then, we present our novel framework, temporal logic Specification-conditioned Decision Transformer (SDT), and explain the rationale behind our specific approach of incorporating STL specification in DT.

% In this section, we provide a general formulation for solving offline safe RL problems through supervised learning. 
% Following this problem formulation, we present our approach to addressing the problem, the temporal logic Specification-conditioned Decision Transformer (SDT).
% Then, we explain the details of our design, focusing on the rationale behind using prefix and suffix robustness values in our method.

% \vspace{-2mm}
\subsection{Offline RL via Supervised Learning}
Motivated by RCSL~\cite{emmons2021rvs, brandfonbrener2022does}, we extend the supervised learning formulation to the offline safe RL setting.
Instead of solving the constrained optimization problem in Eq.~(\ref{eq:safe-rl}), the objective is to find an autoregressive model to simulate sampled trajectories conditioned on both reward and cost:
% \vspace{-4mm}
\begin{equation}
\begin{aligned}
    \max_{\theta} \mathbb{E}_{\mathbf{s} \sim \tau, \tau \sim \mathcal{D}} \Bigl[\log \pi_{\theta} \bigl(a | \mathbf{s}, r(\tau), c(\tau) \bigl) \Bigl] + \mathcal{L}_{\text{reg}}
    % & + regularization
    \label{eq:loss}
\end{aligned}
\end{equation}
% \vspace{-5mm}
where $\mathcal{L}_{\text{reg}}$ is the regularization term; $\mathbf{s} = \{s_{t-K}, ..., s_t\}$ is a sequence of states with a context length $K$; $r(\tau)$ and $c(\tau)$ represent the rewards and costs of the trajectory.
This formulation captures a range of methods, as detailed in Table~\ref{tab:algos}. 
RvS~\cite{emmons2021rvs} uses MLP policy with $K=1$ and uses the $r(\tau)$ function to compute either the average future reward or future goal states. 
DT~\cite{chen2021decision} adopts a transformers-based policy with $K > 1$ and uses $r(\tau)$ to specify return-to-go. 
ODT~\cite{zheng2022online} and QDT~\cite{yamagata2023q} follow similar principles as DT, but ODT incorporates stochastic policies and entropy regularization, and QDT learns Q-functions to relabel return-to-go. 
CDT~\cite{liu2023constrained} extends to offline safe RL, utilizing cost-to-go in the $c(\tau)$ function. 
SaFormer~\cite{zhang2023saformer} estimates the cost-to-go to filter unsafe actions. 
% This framework's ability to encompass diverse policy, reward, and cost functions, both Markovian and non-Markovian, demonstrates its versatility in tackling a broad spectrum of problems. The focus is on non-Markovian policies conditioned on temporal behavior specifications and non-Markovian rewards and costs.
Despite the variety, these existing methods center on  Markovian rewards and costs.
%limiting the potential of the non-Markovian policy (Transformers).
Our approach combines non-Markovian policies (transformers) with non-Markovian costs 
(robustness values of STL specifications).
% \vspace{-2mm}
\begin{table}[ht]
\centering
\LARGE
\renewcommand{\arraystretch}{1.7}
\resizebox{1.0\linewidth}{!}{
\begin{tabular}{ccccc}
% {|c|c|c|c|c|}
% \hline
\Xhline{2pt}
Method                        & Architecture & $r$                                                              & $c$               & $\mathcal{L}_{\text{reg}}$ \\ \hline
RvS-R~\cite{emmons2021rvs}    & MLP          & average reward                                                   & $-$               & $-$                        \\
RvS-G~\cite{emmons2021rvs}    & MLP          & goal state                                                       & $-$               & $-$                        \\ \hline
DT~\cite{chen2021decision}    & Transformers & return-to-go                                                     & $-$               & $-$                        \\
ODT~\cite{zheng2022online}    & Transformers & return-to-go                                                     & $-$               & Entropy                    \\
QDT~\cite{yamagata2023q}      & Transformers & \begin{tabular}[c]{@{}c@{}}relabeled\\[-14pt] return-to-go\end{tabular} & $-$               & $-$                    \\ \hline
CDT~\cite{liu2023constrained} & Transformers & return-to-go                                                     & cost-to-go        & Entropy                    \\
SaFormer~\cite{zhang2023saformer} & Transformers & return-to-go                                                 & \begin{tabular}[c]{@{}c@{}}estimated\\[-14pt] cost-to-go\end{tabular}        & $-$                    \\ \hline
SDT(ours)                     & Transformers & return-go-to                                                     & robustness values & Entropy                    \\ \Xhline{2pt}
% \hline
\end{tabular}
}
% \vspace{-1mm}
\caption{A brief comparison of methods with different architectures conditioned on different types of inputs. $r$ (or $c$): reward (or cost) function.}
\vspace{-3mm}
\label{tab:algos}
\end{table}

\subsection{Specification-Conditioned Decision Transformers}
\label{sec:sdt}

\begin{figure}[ht]
\centering
\includegraphics[width=1.0\linewidth, trim={0 1cm 0 3cm}, clip]{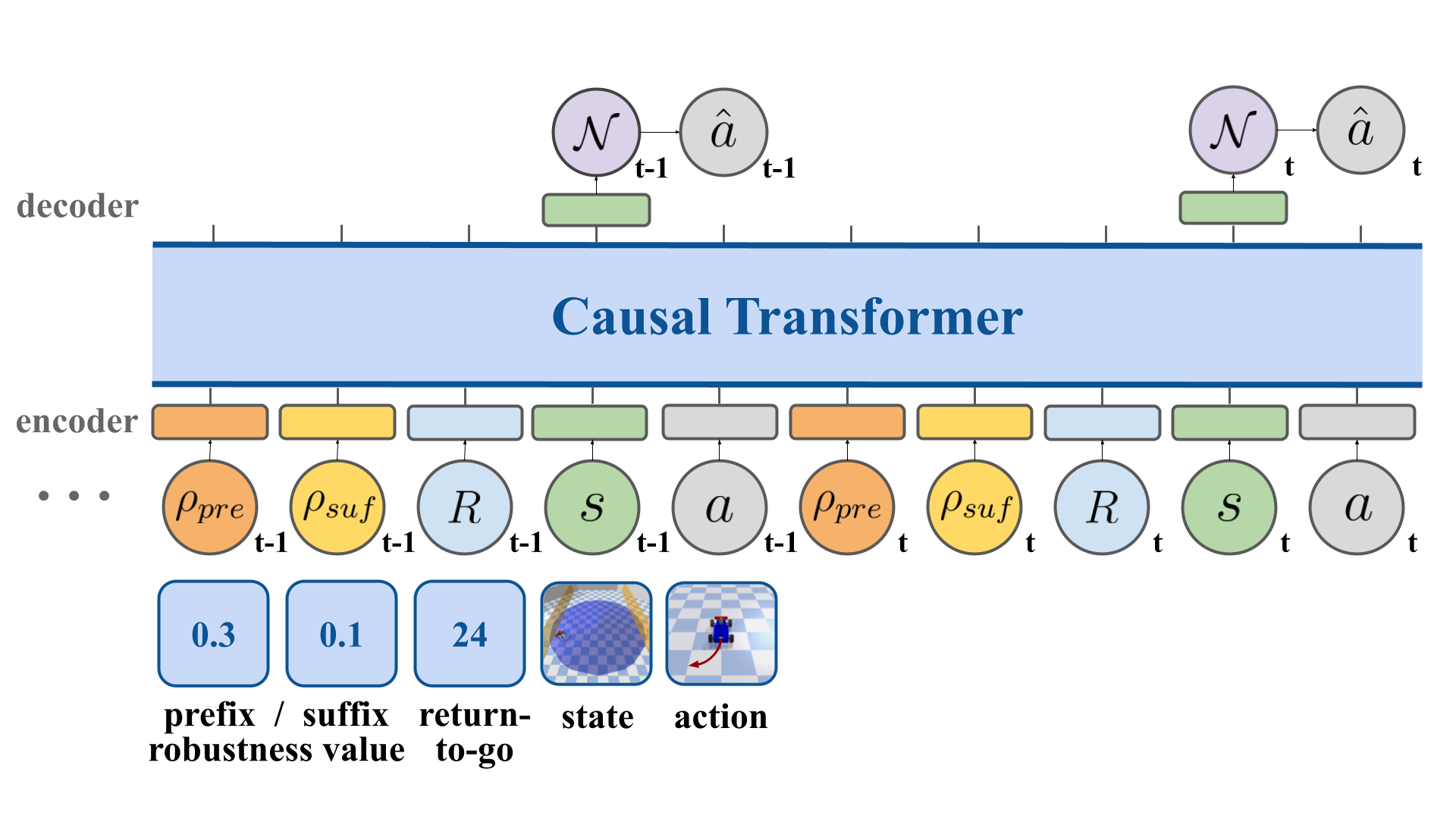}
% \includegraphics[width=.98\linewidth]{figures/sdt-corrected.png}
% \vspace{-2mm}
\caption{The SDT framework. It takes the prefix and suffix robustness values, return-to-go, states, and actions as inputs and predicts the next actions using a Gaussian policy.}
\label{fig:sdt}
% \vspace{-4mm}
\end{figure}

Our framework, named temporal logic Specification-conditioned Decision Transformer (SDT), as illustrated in Figure~\ref{fig:sdt}, extends the Decision Transformer (DT) model~\cite{chen2021decision} by incorporating two extra robustness value tokens, prefix and suffix, as defined in Definition~\ref{def:pre&suf}.
The intuition is to generate actions conditioned on both the reward and the trajectory's robustness in satisfying the specification.
In this work, we focus on safety specifications\footnote{In formal logic parlance, all specifications over finite trajectories are safety specifications. We use the term ``safety" here to differentiate safety constraints from performance goals.} and preserve the reward token to measure performance.
We employ a stochastic policy with entropy regularization, which has been shown to be effective in the literature~\cite{zheng2022online, liu2023constrained}.

% \vspace{-2mm}

% \vspace{-1mm}
\begin{definition}
\label{def:pre&suf}
% (Prefix and suffix robustness value)
(Prefix and suffix robustness value) Given a trajectory $\tau = \{s_1, s_2, ..., s_T\}$ with length $|\tau| = T$ and a STL specification $\phi$, the prefix and suffix robustness value at time-step $t$ are $\rho_{pre}(\tau, t, \phi):=\rho(\tau_{1:t}, 1, \phi)$ and $\rho_{suf}(\tau, t, \phi):=\rho(\tau_{t:T}, 1, \phi)$ respectively.
% Without otherwise statements, we will abbreviate "prefix robustness value" as "prefix" and "suffix robustness value" as "suffix" for simplicity.
For convenience, we will abbreviate “prefix and suffix robustness values” as “prefix” and “suffix" respectively.
\end{definition}
% \vspace{-2mm}

Specifically, the input sequence $\mathbf{o}_t$ for SDT at time-step $t$ includes prefix $\mathbf{P}_{pre} = \{ \rho_{pre}(\tau, t-K, \phi)$, $...$, $\rho_{pre}(\tau, t, \phi) \}$, suffix $\mathbf{P}_{suf} = \{ \rho_{suf}(\tau, t-K, \phi)$, $...$, $\rho_{suf}(\tau, t, \phi) \}$, rewards $\mathbf{R}_{t-K:t}$, states $\mathbf{s}_{t-K:t}$, and actions $\mathbf{a}_{t-K:t}$; the Gaussian policy parameterized by $\theta$ is $\pi_\theta (\cdot | \mathbf{o}_t) = \mathcal{N}\left( \mu_{\theta}(\mathbf{o}_t), \Sigma_\theta(\mathbf{o}_t) \right)$; and the Shannon entropy regularizer is $\mathcal{L}_{\textbf{reg}} = H[\pi_\theta(\cdot | \mathbf{o})]$ with weight $\lambda \in [0, \infty)$.
After plugging them into Eq.~(\ref{eq:loss}), the objective of SDT is:% to optimize:
% \vspace{-5mm}
\begin{equation}
    \max_\theta \quad \mathbb{E}_{\mathbf{o}\sim \tau, \tau \sim \mathcal{D}}  \bigl[ \log \pi_\theta (\mathbf{a} | \mathbf{o}) + \lambda H[\pi_\theta(\cdot | \mathbf{o})] \bigr]
    \label{eq:sdt-loss}
\end{equation}
% \vspace{-6mm}
\textbf{Training and evaluation.}
SDT generally follows the training and evaluation schemes of RCSL~\cite{emmons2021rvs, brandfonbrener2022does}.
The training procedure is as follows: sample a batch of sequences $\{\mathbf{o}, \mathbf{a}\}$ from the offline dataset $\Dcal$, and then compute the loss in Eq. (\ref{eq:sdt-loss}) to optimize the policy $\pi_\theta$ via gradient descent.
In terms of evaluation, the procedure for SDT is presented in Algorithm~\ref{algo:evaluation}.
Updates to the return-to-go and the prefix occur in response to new rewards and state information from the environment.
The target suffix at each time-step is specified instead of autoregressively computed, as detailed in the following section where we justify the usage of the suffix and prefix.

\begin{algorithm}[tb]
% \small
% \footnotesize
\caption{Evaluation procedure for SDT}
{\bfseries Input:} \raggedright trained Transformer policy $\pi_\theta$, STL specification $\phi$, episode length $T$, context length $K$, target reward $R$ and target suffix $\mathbf{P}_{suf}$, environment $env$ \par
% \begin{minipage}{0.9\linewidth}

\begin{algorithmic}[1]
\STATE Get the initial state: $s_1 \leftarrow env.reset()$ and initial action: $a_t \leftarrow 0$ and initial input sequence: $\mathbf{o} \leftarrow \emptyset$.
\FOR{$t=1,..., T$}
\STATE Compute prefix $\rho_{pre} = \rho(s_{1:t}, \phi)$
\STATE Set target suffix $\rho_{suf} = \mathbf{P}_{suf}[t]$
\STATE Construct input sequence $\mathbf{o} \leftarrow \mathbf{o} \cap \{\rho_{suf}, \rho_{pre}, R_t, s_t, a_t\}$
\STATE Get predicted action $\hat{a}_t \sim \pi(\cdot | \mathbf{o}_{t-K+1:t})$
\STATE Execute the action: $s_{t+1}, r_t \leftarrow env.step(\hat{a}_t)$
% \STATE Compute prefix $\rho_{pre}$ based on $s_{1:t}$
\STATE Compute target returns for the next step $R_{t+1} = R_t-r_t$ and set $a_t \leftarrow \hat{a}_t$
% \STATE Append the new token $o_t = \{ \rho_{suf}, \rho_{pre}, R_{t+1}, s_{t+1}, \hat{a}_t \}$ to the sequence $\mathbf{o}$
\ENDFOR
% \end{algorithmic}
\end{algorithmic}
\label{algo:evaluation}
% \vspace{-5mm}
% \end{minipage}
\end{algorithm}
% \vspace{-5mm}
% \end{minipage}

% \vspace{-2mm}
\subsection{Suffix and Prefix Robustness Values}
\label{sec:stl-rob}

\textbf{Suffix robustness value as desired future.}
Recall that given a trajectory $\tau$ of length $T$, the return-to-go (or cost-to-go), at time-step $t$ is computed by $R_t = \sum_{t'=t}^{T} r(s_{t'}, a_{t'})$ (or $C_t = \sum_{t'=t}^{T} c(s_{t'}, a_{t'})$).
During training, they are concatenated with a sequence of states and actions and then fed to DT to optimize Eq.~(\ref{eq:loss}).
\cite{furuta2021generalized} point out that the conditional supervised learning approaches are performing \textit{hindsight information matching}: match the 
output trajectories with future \textit{information statistics} $I(\tau)$ to search for optimal actions.
We introduce the suffix robustness value in Definition~\ref{def:pre&suf}, which can be viewed as a particular form of $I(\tau)$, as an alternative to cost-to-go.
For example, given a trajectory $\tau$ and a simple specification $\phi_1 = \mathbf{F}_{[1, 10]}(\mathbf{s} > 0)$ describing that the states in a trajectory $\tau$ should \textit{eventually} be greater than 0 within the next 10 steps.
The corresponding suffix at time-step $t$ can be computed as $\rho_{suf}(\tau, t, \phi_1) = \max_{t'\in [t+1, t+10]} s_{t'}$.
The suffix can capture the statistics of future states.
Moreover, 
STL suffix is strictly more expressive than cost-to-go (or return-to-go), as one can define a predicate as the sum of cost (or negative sum of reward) and the resulting suffix will be equivalent to negative cost-to-go (or return-to-go), e.g., $\footnotesize \mu_0(s_t):= C_t < 0 \Rightarrow \rho(\tau, t, \mu_0) = -C_t$.
However, challenges arise in training and evaluation due to the following sparsity issue and the updates of STL robustness values.

\textit{Sparsity.}
    The compliance or violation of STL specifications of a trajectory might be determined by a relatively small number of critical states or intervals within the trajectory.
    As highlighted by the quantitative semantics of STL in Eq.~(\ref{eq:stl-rob}), the suffix, or even a sequence of suffix
    % $\{ \rho_{suf}(s_{t:T}, \phi)$, $...$, $\rho_{suf}(s_{t+K:T}, \phi) \}$ 
    can offer sparse information on its corresponding states.
    % $\{ s_t, ..., s_{t+k} \}$.
    % For $\phi_1$ and $\phi_2$, It is clear that $\phi_1$ depends on the minimum state in the trajectory, and $\phi_2$ relies on a specific interval of the trajectory.
    For example, given the STL formula $\phi_1$, there may exist a trajectory $\tau$ whose states do not satisfy $\textbf{s}> 0$ until the end of or near the end of $\tau$. 
    Consequently, a large portion of the suffix robustness values $\{ \rho_{suf}(\tau, 1, \phi), ..., \rho_{suf}(\tau, T, \phi) \}$ equal to a constant.
   In this case, the agent cannot gain meaningful information due to limited feedback on robustness values.
    % The suffix alone suffers to guide the agents to perform desired requirements.
% \end{remark}

% \begin{remark}[Monitoring of STL]
\textit{Updates of STL robustness values.}
% \vspace{1mm}
    The cost-to-go (or return-to-go) can be updated autoregressively when a new cost (or reward) is obtained, i.e., $C_{t+1} = C_{t} - c(s_{t}, a_{t})$ (or $R_{t+1} = R_{t} - r(s_{t}, a_{t})$).
    However, the robustness value does not possess this additive property, i.e., $\rho_{suf}(\tau, t, \phi) \neq \rho_{suf}(\tau, t+1, \phi) - \rho(\tau_{t}, 1, \phi)$, 
    Although the relation of two suffixes $\tau_{t:T}$ and $\tau_{t+1:T}$ of a given trajectory $\tau$ can be expressed as $\rho_{suf}(\tau, t, \phi) = f(\rho_{suf}(\tau, t+1, \phi), \rho(\tau_{t}, 1, \phi))$, $f$ is a complex recursive mapping that depends on the specification, i.e., nested $\min()/\max()$ operations~\cite{donze2013efficient, dokhanchi2014line, deshmukh2017robust}.
    Thus, the target suffix for each time-step has to be set in advance instead of computing the target reward and cost in an autoregressive manner.
    Empirically, we test and validate several different target suffix configurations in Section~\ref{sec:ablation}.

\textbf{Prefix robustness value as achieved past.}
To deal with the sparsity issue of the suffix robustness value, we introduce the prefix robustness value, derived from the states of a trajectory up to time-step $t$, supplementary to the suffix.
The intuition is that different segments of the trajectory, i.e., $\tau_{1:t}$ and $\tau_{t:T}$, can provide complementary information on this trajectory.
While the prefix itself is subject to sparsity, we argue that \textit{the comparison between the prefix and the suffix offers additional information about how robust the action is in terms of specification satisfaction.}
Suppose that the prefix and suffix of a trajectory in the offline dataset at time-step $t$ are $\rho_{pre}$ and $\rho_{suf}$.
When both $\rho_{pre}$ and $\rho_{suf}$ are positive, it indicates safety in both previous and future actions.
Conversely, negative values for both suggest unsafe actions throughout the trajectory.
A positive $\rho_{pre}$ and a negative $\rho_{suf}$ implies that while past actions have been safe, a future action at a certain time-step $t'$ will be unsafe.
In contrast, a negative $\rho_{pre}$ and a positive $\rho_{suf}$ indicate unsafe past actions but safe future actions.
These insights can extend beyond the context length, aiding the policy in inferring the safety of both past and future states and actions even if they fall outside the current context, which cannot be achieved by cost-to-go (or return-to-go). 
Therefore, the combination of the prefix and suffix addresses the sparsity issue by providing comprehensive information about the trajectory.
The ablation study in Section~\ref{sec:ablation} shows that including both the prefix and suffix token is not only crucial for our SDT to learn a safe and high-reward policy but also can improve the performance of the standard DT framework, which uses only return-to-go.

\vspace{-2mm}
\section{Experiment}
\label{sec:exp}

% \td{
% \begin{itemize}
%     % \item add RvS-RC as another baseline. 
%     \item goal-conditioned tasks
%     % \item SDT prefix only, SDT suffix only (Ball-Circle, Drone-Circle, Ant-Run, Ball-Run, Drone-Run)
%     % \item SDT and RvS-R$\rho$ alignment results (Ball-Circle, Drone-Circle, Ant-Run, Ball-Run, Drone-Run)
%     % \item SDT evaluation with different suffix
% \end{itemize}}
We use the following environments and STL specifications to evaluate SDT and baseline approaches and aim to answer the following questions:
\begin{itemize}
% \begin{enumerate}
    \vspace{-2mm}
    \setlength{\itemsep}{0.1em}
    \item Can SDT learn policies that satisfy a given STL specification from offline datasets? 
    \item Can SDT align with different target suffixes? 
    \item How important are the prefix and suffix inputs in SDT?
    \item What is the influence of different target suffix configurations on the performance of SDT?
    \item Is SDT robust to rescaling individual predicates?
    \vspace{-1mm}
\end{itemize}
\textbf{Environments.}
% We use several robot locomotion tasks that are commonly used in previous works \citep{achiam2017constrained, chow2019lyapunov}. 
The \texttt{Bullet-Safety-gym}~\cite{gronauer2022bullet} is a public benchmark that includes a variety of robot locomotion tasks commonly used in previous works~\cite{achiam2017constrained, chow2019lyapunov}.
We consider three specific environments, \texttt{Run}, \texttt{Circle}, and \texttt{Reach}, where different types of robots, \texttt{Ball, Car, Drone}, and \texttt{Ant}, are trained. 
In the \texttt{Run} environment, agents earn rewards for achieving high speeds between two boundaries but incur penalties if they cross the boundaries or exceed an agent-specific velocity threshold.
In the \texttt{Circle} environment, agents are rewarded for moving in a circular pattern but are constrained within a safe region smaller than the radius of the target circle.
In the \texttt{Reach} environment, besides performing the same task as in the \texttt{Circle} environment, agents have an additional task of reaching goals in sequence.
This setup of rewards and costs creates a dual influence on the agents' behaviors, where rewards motivate specific actions and costs act as deterrents for those actions.
More details of the environments can be found in Appendix~\ref{app:envs}.

% \vspace{-1mm}
\textbf{Temporal behaviors}. 
To evaluate the capability of SDT to satisfy temporal requirements, we consider the following STL specifications.
\begin{align}
    & \begin{aligned}
        \phi_{\text{run}} = \mathbf{G} \Bigl( \psi_{\text{bndry}} \land \bigl( \neg \psi_{\text{vel}} \Rightarrow \mathbf{F}_{[1, 5]}\psi_{\text{vel}} \bigl)  \Bigl)
        \label{eq:spec-run}
    \end{aligned} \\
    & \begin{aligned}
        & \phi_{\text{circle}} = \mathbf{G} \Bigl( \neg \psi_{\text{bndry}} \Rightarrow \mathbf{F}_{[1, 5]}\psi_{\text{bndry}} \Bigl)
        \label{eq:spec-circle}
    \end{aligned}
    \\
    & \begin{aligned}
        \phi_{\text{reach}} = \phi_{\text{circle}} \land \mathbf{F} \Bigl( \neg \psi_{\text{goalB}} \mathbf{U} \psi_{\text{goalA}} \Bigl)
    \end{aligned}
    % & \begin{aligned}
    %     & \phi_{\text{reach}} = \mathbf{G} \Bigl( \neg \psi_{\text{bndry}} \Rightarrow \mathbf{F}_{[1, 5]}\psi_{\text{bndry}} \Bigl) \land \mathbf{F} \Bigl( \neg \psi_{\text{goalB}} \mathbf{U} \psi_{\text{goalA}} \Bigl)
    %     \label{eq:spec-reach}
    % \end{aligned}
\end{align}
where $\psi_{\text{bndry}}: s_t < d_{\text{lim}}$ is the predicate for staying within the safety boundary denoted as $d_{\text{lim}}$, 
$\psi_{\text{vel}}: s_t < v_{\text{lim}}$ is the predicate for maintaining a safe velocity denoted as $v_{\text{lim}}$, 
and $\psi_{\text{goal}}$\footnote{$\psi_{\text{goal}}$ is defined as $-s_t < d - x \land s_t < x + d \land -s_t < d - y \land s_t < y + d$ with $\text{goal} = [x, y]$ and $\rho(\tau, t, \psi_{\text{goal}}) = \min(-s_t+x-d, s_t-x-d, -s_t+y-d, s_t-y-d)$.} is the specification for reaching a small square region (since the predicates are linear) near a goal position $s_{\text{goal}}$.
Note that the state of the agent $s_t$ contains the position and velocity information.
In the \texttt{Run} environment, the agent is required to always stay between the boundaries, and if it exceeds the velocity threshold, it should slow down within the next 5 steps.
In the \texttt{Circle} environment, the agent must leave the unsafe region within the next 5 steps once it enters the unsafe region.
In the \texttt{Reach} environment, the agent must leave unsafe regions within a certain number of steps but also reach two goals in sequence, both located in the safe region, at least once.
Although the specifications relax the original cost as it allows agents to enter the unsafe region as long as they can re-enter the safe region, they completely change the Markovian property of the original cost, making it challenging to learn safe policies.
The corresponding robustness values of the prefix and suffix at time-step $t$ can be calculated as:
% \begin{equation}

{\small
\vspace{-5mm}
\begin{align}
    &
    \begin{aligned}
        \small
        \rho(\tau, t, & \phi_{\text{run}}) = 
        \min_{t'\in[t_1, t_2]} \biggl( \min \Bigl( d_{\text{lim}}-s_{t'} , \\[-6pt]
        & \max \bigl( (v_{\text{lim}} - s_{t'}), 
        \max_{t'' \in [t'+1, t'+5]}(v_{\text{lim}} - s_{t''}) \bigl) \Bigl) \biggl)
    \end{aligned} \\
    &
    \begin{aligned}
        \small
        \rho(\tau, t, \phi_{\text{circle}}) = 
        & \min_{t'\in[t_1, t_2]} \Bigl(
        \max \bigl( (d_{\text{lim}} - s_{t'}), \\[-6pt]
        & \max_{t'' \in [t'+1, t'+5]}(d_{\text{lim}} - s_{t''}) \bigl) \Bigl)
    \end{aligned} \\
    &
    \begin{aligned}
        \small
        \rho(\tau, & t, \phi_{\text{reach}}) = \min\biggl(\rho(\tau, t, \phi_{\text{circle}}),
        \max_{t' \in [t_1, t_2]} \Bigl(\max_{t'' \in [t', t_2]} \\
        & \min \bigl( \rho(\tau, t'', \psi_{\text{goalA}}), \min_{t''' \in [t', t'']} \rho(\tau, t''', \neg \psi_{\text{goalB}}) \bigl) \Bigl)\biggl)
    \end{aligned}
\end{align}
% \end{equation}
}
\vspace{-3mm}

where $I = [t_1, t_2]$ is the interval.
By setting $I$ to be $[1, t] $ or $[t, T]$, we can obtain the prefix and suffix, respectively.

% \vspace{-6mm}
% We can see that the robustness value consists of nested $\min()/\max()$ functions to describe temporal requirements and would be difficult to design by hand but can be generated from the quantitative semantics of STL.
% In \texttt{Circle} environments, we define the STL specification as:
% \begin{equation}
%     \phi_{circle} = \mathbf{G} \Bigl( \neg (|x_t| \leq x_{lim}) \Rightarrow \mathbf{F}_{[1, 5]}(|x_t| \leq x_{lim})  \Bigl)
%     \label{eq:spec-circle}
% \end{equation}
% where $|x_t| \leq x_{lim} \equiv (x_t \leq x_{lim} \land x_t)$

% In \texttt{Run} tasks, we define the STL specification as:
% \begin{equation}
%     \begin{aligned}
%         \phi_{run} = \mathbf{G} \Bigl( & (|y_t| \leq y_{lim}) \land \\
%         & \bigl( \neg (v_t \leq v_{lim}) \Rightarrow \bigl(\mathbf{F}_{[1, 5]}(v_t \leq v_{lim}) \bigl)  \Bigl)
%     \end{aligned}
%     \label{eq:spec-run}
% \end{equation}

% \vspace{-1mm}
\textbf{Offline dataset.}
We use the dataset from \texttt{DSRL}~\cite{liu2023datasets}, a comprehensive benchmark specialized for offline safe RL. 
Note that although this dataset contains the behaviors defined in the STL specifications, it is not designed for tasks with temporal constraints. 
For \texttt{Run} and \texttt{Circle} environments, to be consistent with the STL specifications, we define a new cost function to relabel the original cost $c$ in the dataset:

\resizebox{\linewidth}{!}{%
$
c_t' = 
\begin{cases}
    1 & \text{if } c_{p_t} = 1 \text{ or } [c_{v_{t-5}}, \dots, c_{v_{t-1}}] = \mathbf{1} \text{ for \texttt{Run} envs} \\
    1 & \text{if } [c_{p_{t-5}}, \dots, c_{p_{t-1}}] = \mathbf{1} \text{ for \texttt{Circle} envs} \\
    0 & \text{otherwise}
\end{cases}
$
}

where $c_t'$ is the relabeled cost at time-step $t$ and $c_{p_t}$ and $c_{v_t}$ denote the costs related to the position and velocity of agents, respectively, in the original dataset. 
The cumulative relabeled cost captures the violations against an STL specification, i.e., $C'_t = \sum_1^T c'_t = 0 \Leftrightarrow \rho(\tau_{1:T}, T, \phi) > 0$ for a trajectory $\tau = \{ s_t, a_t, r_t, c'_t \}_{t=1}^T$.
% which means that the 
The relabeled cost-reward plots and suffix-reward plots are presented in Appendix~\ref{app:data-vis}.
For \texttt{Reach} environment, it is challenging to devise a cost function to capture the constraint precisely, i.e. relabel the cost in the offline dataset.
Thus, we only train and evaluate the baselines that do not require costs. 
Moreover, it highlights the benefits of STL and the versatility of our method, which is applicable in broader scenarios, including both Markovian and non-Markovian constraints.

\textbf{Metrics.} 
Our evaluation metrics include (i) normalized cumulative reward, (ii) cumulative relabeled cost, which are consistent with the offline RL literature~\cite{fu2020d4rl, liu2023datasets}, and (iii) the satisfaction rate that indicates the ratio of episodes in which the STL specification is satisfied within a maximum number of time-steps.
% \begin{itemize}
%     \item Normalized reward return: 
% \end{itemize}
We use the actual cumulative relabeled cost instead of the normalized one since $d = 0$.
The normalized cumulative reward is:
% computed by:
% \vspace{-4mm}
$$
    R_{\text{normalized}} = \frac{R_\pi - r_{\text{min}}(\Dcal)}{r_{\text{max}}(\Dcal)-r_{\text{min}}(\Dcal)}
$$
% \vspace{-3mm}
where $R_\pi$ is the evaluated cumulative reward of policy $\pi$ and $r_{\text{max}}(\Dcal)$ and $r_{\text{min}}(\Dcal)$ are the maximum cumulative reward and the minimum cumulative reward of the trajectories that satisfy the cost threshold in dataset $\Dcal$.
For convenience, we will abbreviate “normalized cumulative reward” as “reward” and “cumulative relabeled cost” as “cost”.

\textbf{Baselines.}
We present our results by comparing SDT with two categories of baselines: (i) constrained optimization methods and (ii) conditioned RL methods.
% Specifically, the implementation of these baselines is directed at addressing the problems of satisfying temporal specifications within the context of sequential decision-making tasks.
% The offline Safe RL baselines generally solve a constrained optimization problem of CMDP.

% \vspace{-1mm}
%\skip
For the constrained optimization RL baselines, we consider two Lagrangian-based methods that use adaptive PID-based Lagrangian multipliers~\cite{stooke2020responsive} to penalize constraint violations: BCQ-Lagrangian (BCQ-Lag) and BEAR-Lagrangian (BEAR-Lag), which is built upon BCQ~\cite{fujimoto2018off} and BEAR~\cite{kumar2019stabilizing}, respectively.
We also include CPQ~\cite{xu2022constraints}, designed to learn safe policy by conservative cost estimations, and CoptiDICE~\cite{lee2022coptidice}, which learns safe policies via stationary distribution correction.

% \vspace{-1mm}
For the conditioned RL baselines, we consider CDT~\cite{liu2023constrained} and construct two variants of the reward-conditioned RvS-R~\cite{emmons2021rvs}: RvS-RC with an additional cost token and RvS-R$\rho$ with two additional prefix and suffix tokens.
We also include a Behavior Cloning baseline (BC-Safe) that only uses trajectories that satisfy the specifications to train the policy. 
% This measures whether each method performs effective RL or simply copies the data.\li{need to discuss}
Following the safe RL setting criteria~\cite{ray2019benchmarking}, we prioritize safety: a policy that maintains a high satisfaction rate is preferred over ones that do not.
The methods with a higher satisfaction rate than BC-safe are considered safe since strict zero-constraint violation is difficult if not impossible in a model-free setting.
% \li{point out model-free earlier?}.
We use a small, fixed target suffix above zero at each time-step for methods trained on robustness values, except for the ablation study on SDT about target suffix configuration.

% \textbf{Questions.} 
% We design experiments and corresponding ablation studies to evaluate the proposed approach and empirically answer the following questions
% \begin{itemize}
%     \item[Q1] Can SDT learn policies that satisfy the specification from offline datasets? 
%     \item[Q2] Is SDT adaptable to different target suffix thresholds? 
%     \item[Q3] What is the importance of the prefix and suffix in SDT?
%     \item[Q4] How does the target suffix affect SDT's performance?
% \end{itemize}
% 1) can SDT learn policies that satisfy the specification from offline datasets? 
% 2) is SDT adaptable to different target suffix thresholds? 
% 3) what is the importance of the prefix and suffix in SDT?  
% 4) how does the target suffix affect the performance of SDT?

\begin{table*}[ht]
\centering
\renewcommand{\arraystretch}{1.4}
% \Large
\small
% \resizebox{1.\textwidth}{!}{
% \begin{tabular}{c|ccc|ccc}
% \begin{tabular}{ccccccc}
\begin{tabular}{ccccccccc}
% \hline
% \toprule
\Xhline{1.5pt}

                          & \multicolumn{3}{c}{Run-Average}                                                                                      & \multicolumn{3}{c}{Circle-Average}                                                                                   & \multicolumn{2}{c}{Reach-Average}                   \\
\multirow{-2}{*}{Methods} & Reward $\uparrow$                    & Cost $\downarrow$                      & Rate $\uparrow$                      & Reward $\uparrow$                    & Cost $\downarrow$                      & Rate $\uparrow$                      & Reward $\uparrow$        & Rate $\uparrow$          \\ \hline
SDT(ours)                 & \textbf{0.96$\pm$0.01}               & 0.33$\pm$2.04                          & \textbf{0.97$\pm$0.04}               & 0.89$\pm$0.08                        & \textbf{0.94$\pm$3.73}                 & \textbf{0.86$\pm$0.1}                & \textbf{0.75 $\pm$ 0.13} & \textbf{0.76 $\pm$ 0.19} \\
RvS-R$\rho$               & 0.79$\pm$0.28                        & 8.89$\pm$20.28                         & 0.79$\pm$0.32                        & 0.81$\pm$0.11                        & 1.09$\pm$3.19                          & 0.8$\pm$0.14                         & 0.55 $\pm$ 0.25          & 0.65 $\pm$ 0.22          \\ \hline
CDT                       & \textbf{0.96$\pm$0.03}               & \textbf{0.32$\pm$1.62}                 & 0.92$\pm$0.12                        & \textbf{0.92$\pm$0.03}               & 2.23$\pm$4.2                           & 0.49$\pm$0.09                        & -                        & -                        \\
RvS-RC                    & {\color[HTML]{656565} 1.27$\pm$0.63} & {\color[HTML]{656565} 20.34$\pm$26.69} & {\color[HTML]{656565} 0.53$\pm$0.41} & 0.75$\pm$0.26                        & 9.65$\pm$29.64                         & 0.62$\pm$0.1                         & -                        & -                        \\
BC-safe                   & 0.78$\pm$0.29                        & 3.18$\pm$8.5                           & 0.78$\pm$0.25                        & 0.62$\pm$0.28                        & 5.75$\pm$10.51                         & 0.49$\pm$0.13                        & 0.58 $\pm$ 0.3           & 0.34 $\pm$ 0.15          \\ \hline
BCQ-Lag                   & {\color[HTML]{656565} 0.9$\pm$0.3}   & {\color[HTML]{656565} 30.08$\pm$39.49} & {\color[HTML]{656565} 0.3$\pm$0.36}  & {\color[HTML]{656565} 0.99$\pm$0.34} & {\color[HTML]{656565} 33.32$\pm$21.25} & {\color[HTML]{656565} 0.08$\pm$0.13} & -                        & -                        \\
BEAR-Lag                  & {\color[HTML]{656565} 0.85$\pm$1.16} & {\color[HTML]{656565} 60.3$\pm$46.47}  & {\color[HTML]{656565} 0.33$\pm$0.47} & {\color[HTML]{656565} 1.07$\pm$0.18} & {\color[HTML]{656565} 33.2$\pm$18.3}   & {\color[HTML]{656565} 0.0$\pm$0.0}   & -                        & -                        \\
CPQ                       & {\color[HTML]{656565} 1.16$\pm$1.33} & {\color[HTML]{656565} 50.29$\pm$48.38} & {\color[HTML]{656565} 0.33$\pm$0.47} & 0.59$\pm$0.35                        & 5.73$\pm$18.94                         & 0.72$\pm$0.36                        & -                        & -                        \\
COptiDICE                 & {\color[HTML]{656565} 0.9$\pm$0.13}  & {\color[HTML]{656565} 21.83$\pm$27.68} & {\color[HTML]{656565} 0.41$\pm$0.46} & {\color[HTML]{656565} 0.66$\pm$0.13} & {\color[HTML]{656565} 12.41$\pm$17.57} & {\color[HTML]{656565} 0.17$\pm$0.13} & -                        & -                        \\
\Xhline{1.5pt}
\end{tabular}
% }
\caption{Evaluation results of reward, cost, and satisfaction rate. 
$\uparrow$: the higher the reward, the better.
$\downarrow$: the lower the cost (closer to 0), the better.
Agents with a higher satisfaction rate than BC-safe are considered safe. 
{\color[HTML]{656565} Gray}: unsafe agents.
\textbf{Bold}: the best (highest reward, highest specification satisfaction rate, and lowest cost) in the respective metric.
The satisfaction rates of the trajectories in the offline dataset are $20.0\%$, $8.52\%$, and $3.47\%$ for \texttt{Run}, \texttt{Circle}, and \texttt{Reach} environments.
For \texttt{Reach} environment, it is challenging to devise a cost function to capture the constraint precisely, i.e. relabel the cost in the offline dataset.
Thus, we only train and evaluate the baselines that use robustness values for this environment. 
}
% Each value is averaged over 20 episodes and 3 seeds.
% \vspace{-2mm}
\label{tab:temporal-exp-results}
\end{table*}

% \vspace{-1mm}
\subsection{Can SDT learn specification-satisfying policies?}
% \vspace{-1mm}
\label{sec:cost-fails}
The evaluation results for different trained policies are presented in Table \ref{tab:temporal-exp-results}.% and Table~\ref{tab:reach-results}. 
These results reflect the average performance, with each plot aggregating data from 3 random seeds and 20 trajectories per seed. 
Complete results for each environment are included in Appendix\ref{app:exp-results} due to page limit.
\textit{Our method demonstrates the best performance compared to the baselines in terms of reaching the highest rate of specification satisfaction and lowest costs}. Also, it produces comparable or higher rewards compared to the safe baselines, indicating that SDT performs effective learning.

The results of CDT and RvS-RC show that merely relabeling the cost to reflect STL specifications cannot obtain safe policies. 
Such cost function introduces a level of stochasticity~\cite{paster2022you}, where taking the same action in the same states can lead to inconsistent cost due to the temporal constraint.
Note that in the \texttt{Run} environments, RvS-RC even outperforms the best safe trajectory's reward in the dataset; however, the satisfaction rate is low and the cost is high.
This pattern, observed in most of the baselines, where a high reward often correlates with a high cost, underscores the inherent trade-off between rewards and costs~\cite{liu2022robustness, li2023guided}.
RvS-R$\rho$ shows improved results over RvS-RC, as it benefits from training with the prefix and suffix.
Our method still outperforms it due to the sequential structure of transformers, which is adept at capturing the temporal properties in robustness values, states, and actions.

% \vspace{-1mm}
The Lagrangian-based baselines, BCQ-Lag and BEAR-Lag, as well as CPQ and COptiDICE methods that are tailored for offline safe RL, struggle to ensure safety in most environments.
This indicates that directly applying widely used safe RL techniques to satisfy temporal specifications does not work well.
A key factor for the poor performance is the non-Markovian nature of the relabeled cost since these methods are designed to estimate value functions based on Markovian costs.
As the reward is unchanged in the dataset, the baselines manage to secure high rewards.
We can also notice that the reward of CPQ in the \texttt{Run} environments and the reward of BEAR-Lag in the \texttt{Circle} environments exceed the maximum reward of the safe trajectories in the dataset, which shows the stitching property of the Q-learning-based methods~\cite{yamagata2023q, wang2023critic}.
The poor safety performance of the baselines highlights the challenges posed by the temporal constraints. In comparison, our proposed SDT method successfully learns both safe and rewarding policies in these challenging environments.

% \begin{table}[ht]
% \centering
% \small
% \renewcommand{\arraystretch}{1.4}
% \begin{tabular}{ccc}
% \Xhline{1.5pt}
% \multirow{2}{*}{Methods} & \multicolumn{2}{c}{Reach-Average}                   \\ 
%                          & Reward $\uparrow$        & Rate $\uparrow$          \\ \hline
% SDT(ours)                & \textbf{0.75 $\pm$ 0.13} & \textbf{0.76 $\pm$ 0.19} \\
% RvS-R$\rho$              & 0.55 $\pm$ 0.25          & 0.65 $\pm$ 0.22          \\
% BC-safe                  & 0.58 $\pm$ 0.3           & 0.34 $\pm$ 0.15          \\
% \Xhline{1.5pt}
% \end{tabular}
% \caption{
% Evaluation results of reward and satisfaction rate for \texttt{Reach} environment. }
% \label{tab:reach-results}
% \end{table}

\begin{figure*}[ht]
\centering
\includegraphics[width=1.0\linewidth]{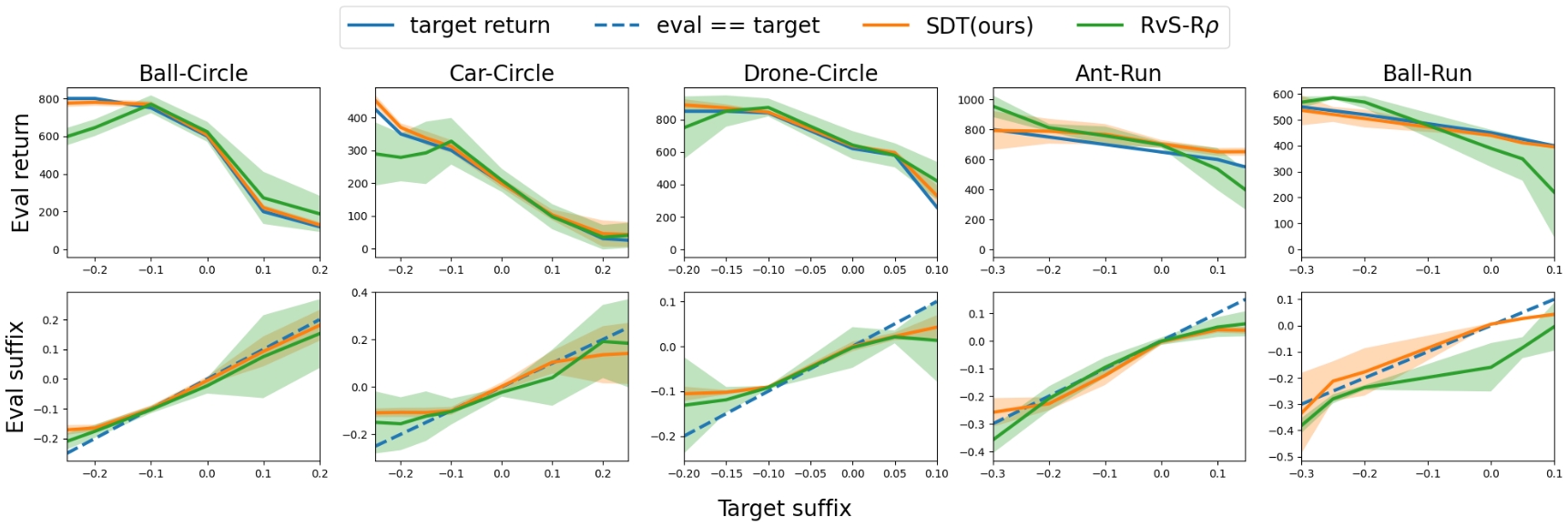}
% \vspace{-3mm}
\caption{Results of alignment with different target suffixes. 
The top-row plots show the evaluated reward and the bottom-row plots show the evaluated suffix.
The solid line and the light shade area represent the mean and mean $\pm$ standard deviation.
}
\vspace{-2mm}
\label{fig:alignmnet}
\end{figure*}

% \vspace{-2mm}
\subsection{Can SDT adapt to different target suffixes?}
\label{sec:alignment}
% \vspace{-1mm}
One of the critical attributes of transformers is the ability to align with the variables on which they are conditioned.
To illustrate this, we vary the target reward and target suffix for evaluation rollouts and obtain the results in Figure~\ref{fig:alignmnet}.
% The target reward is selected as 
It is evident that the baselines introduced previously, except RvS, lack this capability because they depend on a constant and pre-defined threshold to solve a constrained optimization problem and need re-training for adaptation to new constraints.
Therefore, our comparison is focused on SDT and RvS-R$\rho$.
\textit{The results show a strong correlation between the actual and target suffixes for SDT.}
We can also observe that the actual suffix of SDT is above the dashed threshold line when the target suffix is less than zero in most of the environments, which means that SDT can achieve safer actions even under conditions that demand otherwise. 
While there is a saturation point in the curves at specific target suffixes,
% , e.g., $-0.1$ in \texttt{Car-Circle} and \texttt{Drone-Circle}, 
SDT consistently upholds safety even when extrapolating over target suffixes and rewards that represent conflicting objectives.
For example, setting both a high target reward and a high target suffix is unreachable since the former encourages agents to stay close to the safety boundary while the latter advises against it. 
In contrast, RvS-R$\rho$ struggles to meet the target reward when the target suffix is either large or small, further showcasing the strong alignment capabilities of our method.
% in these evaluations.

% \vspace{-1mm}
\subsection{Ablation studies}
\label{sec:ablation}
\begin{figure*}[ht]
\centering
\includegraphics[width=1.0\linewidth]{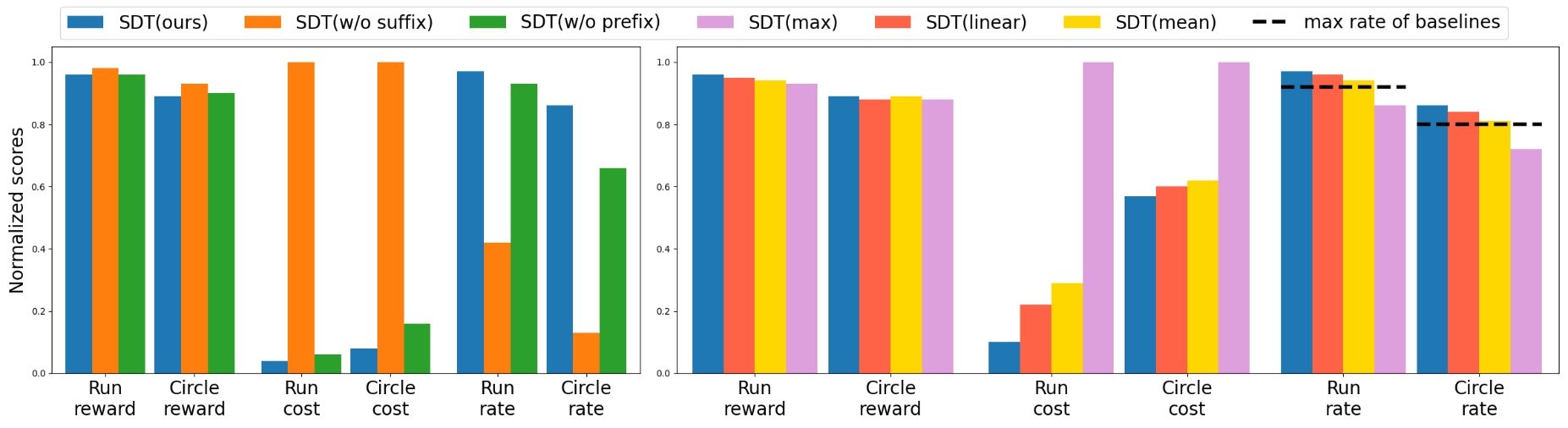}
\vspace{-5mm}
\caption{Ablation study. Left: the effect of the prefix and the suffix. 
Right: influence of different target suffix configurations.
}
\vspace{-2mm}
\label{fig:ablation}
\end{figure*}

\textbf{How important are the prefix and suffix inputs in SDT?}
% \textbf{The prefix and suffix are necessary and vital for SDT.}
To assess the influence of the prefix and suffix, we conduct experiments by removing each of them from SDT.
The left plot in Figure \ref{fig:ablation} shows the averaged performance of SDT, with the cost normalized against the highest cost among the corresponding experiments.
The results indicate that both the prefix and suffix are essential for good performance, as their removal leads to noticeable drops in safety and performance.
In addition, the suffix has a more pronounced effect on the safety performance than the prefix, evidenced by a significant reduction in the satisfaction rate when the suffix is excluded since the prefix lacks hindsight information.

% \textbf{Why is explicit input of the prefix and suffix beneficial?}
Following the insights in Section~\ref{sec:stl-rob} that the relation between prefix and suffix can provide additional information, we re-assess the efficacy of DT with an additional reward prefix token $R_{pre} = \sum_{i=1}^{t} r_i$ while keeping all the other aspects unchanged.
More details about the experiments of DT can be found in Appendix~\ref{app:dt-d4rl}.
From Figure~\ref{fig:dt}, we can see that DT with reward prefix achieves higher rewards than the standard DT.
% , which indicates that explicit input of the prefix and suffix is beneficial.
% \begin{remark}[Explicit versus implicit inputs]
% \vspace{-1mm}
Although the reward prefix is implicitly embedded in the return-to-go, i.e., given the return-to-go of a trajectory at each time-step, we can fully recover the reward prefix, 
explicitly supplying the past information to transformers shows advantageous for policy learning.
On the other hand, we cannot extract prefix from suffix in general for STL robustness values, thus necessitating explicit inputs in SDT.
Recent studies have explored how transformers process explicit inputs versus implicit inputs, with varying conclusions as it largely depends on the specific nature of the data and the task~\cite{zhao2019explicit, chu2021conditional, gui2021kat}.
In our opinion, given enough capacity and data, transformers can learn effectively regardless of the input type, but in practice, transformers tend to learn better with explicit inputs.
\begin{figure}[h]
\centering
% \vspace{-2mm}
\includegraphics[width=1.0\linewidth]{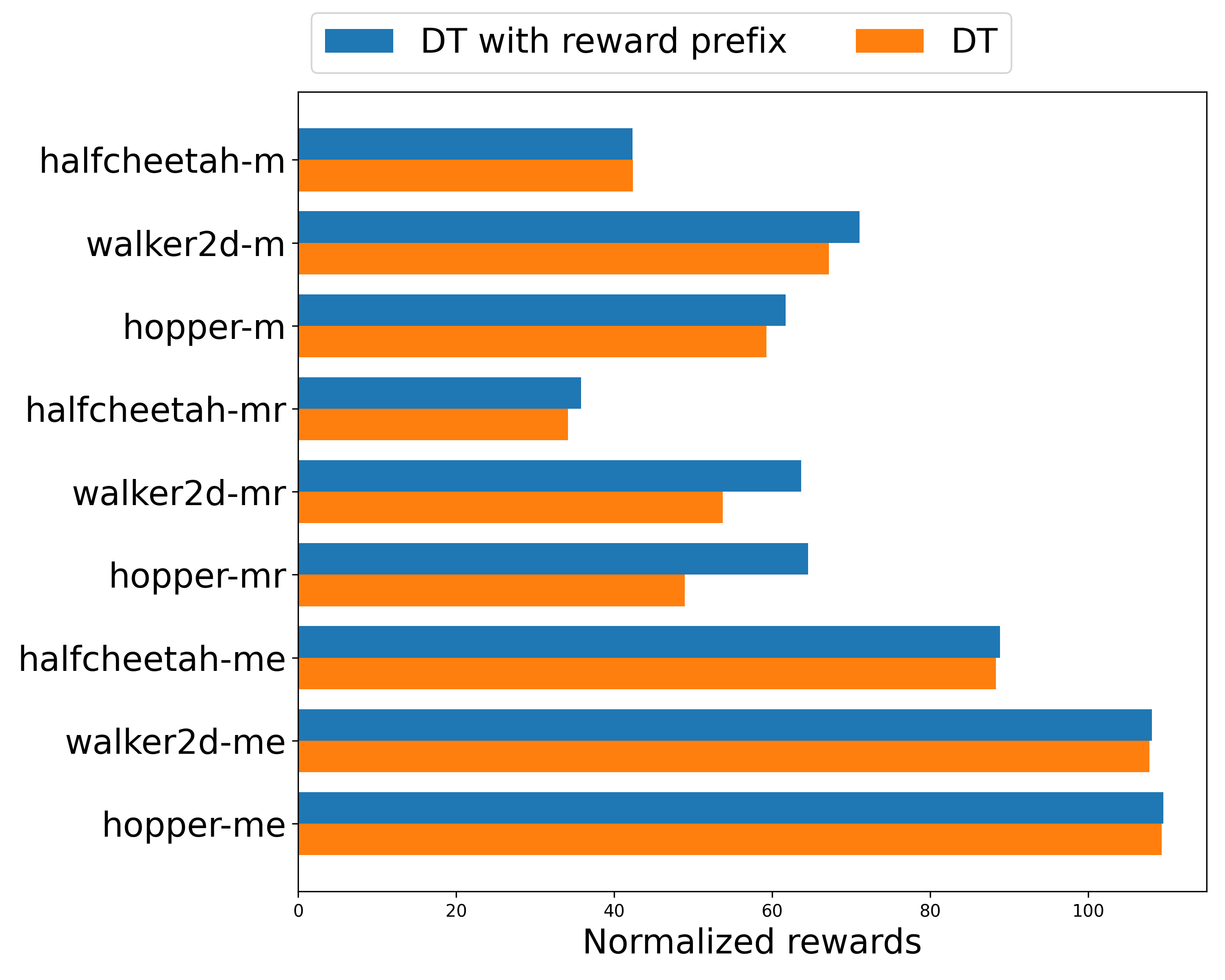}
% \vspace{-5mm}
\caption{Evaluation results of normalized rewards in \texttt{D4RL} Gym environments. 
Each value is averaged over 3 seeds. 
m: medium, mr: medium-replay, me: medium-expert.
}
\vspace{-5mm}
\label{fig:dt}
\end{figure}

% \textbf{SDT is robust to different target suffix configurations.}
\textbf{How do different target suffix configurations influence the performance of SDT?}
% \textbf{Fixed target suffix works best for SDT.}
As mentioned in Section~\ref{sec:sdt}, the suffix offers versatility in evaluations, as we can specify the target suffix for each time-step of the trajectory instead of the auto-regressive manner of return-to-go (or cost-to-go). 
We test SDT using the same target reward but different target suffix configurations: i) SDT(linear), where the target suffix linearly increases each step until it matches the fixed target; ii)  SDT(mean), employing the average suffix from safe trajectories in the dataset; iii) SDT(max), using the maximum suffix from these trajectories.
SDT(linear) is selected as the suffix is monotonically increasing for trajectories in the dataset due to the $\mathbf{G}$ operation in the STL specification in Eq.~(\ref{eq:spec-run}) and (\ref{eq:spec-circle}). 
SDT(mean) aims to mimic the average performance of the safe trajectories, whereas SDT(max) is expected to behave conservatively by emulating the trajectory with the highest suffix.
Details of the prefix and the suffix for dataset trajectories are in Appendix~\ref{app:data-vis}.
As shown in Figure~\ref{fig:ablation}, we can observe that SDT(ours) achieves the best safety performance with a fixed target suffix.
Other target suffix configurations lead to reduced satisfaction rates and increased costs because: i) a low target suffix makes the agent more aggressive, i.e., SDT(linear), and ii) conflicting objectives of achieving a high target suffix and high target reward at the same time, i.e., SDT(mean) and SDT(max). However, it is worth noting that these suffix configurations still outperform the baselines in terms of satisfaction rate.

\begin{table}[ht]
\centering
\small
\renewcommand{\arraystretch}{1.3}
\begin{tabular}{ccc}
\Xhline{1.5pt}
Methods                           & Reward            & Rate            \\ \hline
SDT($\alpha_b=1$, $\alpha_v=1$)   & 0.96 $\pm$ 0.01   & 0.97 $\pm$ 0.04 \\
SDT($\alpha_b=1$, $\alpha_v=10$)  & 0.95 $\pm$ 0.02   & 0.97 $\pm$ 0.06 \\
SDT($\alpha_b=1$, $\alpha_v=100$) & 0.95 $\pm$ 0.03   & 0.99 $\pm$ 0.02 \\
SDT($\alpha_b=10$, $\alpha_v=1$)  & 0.95 $\pm$ 0.02   & 0.96 $\pm$ 0.07 \\
SDT($\alpha_b=100$, $\alpha_v=1$) & 0.93 $\pm$ 0.03   & 0.98 $\pm$ 0.03 \\
\Xhline{1.5pt}
\end{tabular}
\caption{
Evaluation results of reward and satisfaction rate with varying scaling factors. }
\label{tab:scale-predicates}
\end{table}

\textbf{Is SDT robust to rescaling individual predicates?}
\label{sec:scale-predicates}
Though scaling individual predicates does not change the sign of a specification, in practice, it is possible that certain predicates are dominated if they are scaled down, leading to a violation of the specification.
To evaluate the rescaling effect, we incorporate two scaling factors into Eq.~(\ref{eq:spec-run}):
\begin{equation}
    \phi_{\text{run}} = \mathbf{G} \Bigl( \alpha_b \psi_{\text{bndry}} \land \bigl( \neg \alpha_v \psi_{\text{vel}} \Rightarrow \mathbf{F}_{[1, 5]} \alpha_v \psi_{\text{vel}} \bigl)  \Bigl)
    \label{eq:spec-run-scale}
\end{equation}
We explore various pairings of $\alpha_b$ and $\alpha_v$, and the results are shown in Table~\ref{tab:scale-predicates}.
Our findings demonstrate that modifying the scale of the predicates has a negligible impact on both the task performance and the rate of property satisfaction, which indicates that our method is robust to changes in predicate scaling.

\vspace{-1mm}
\section{Conclusion}
\vspace{-1mm}
\label{sec:conclusion}

We study the offline safe RL problem through the lens of supervised learning and point out the unique challenges associated with enforcing temporal constraints.
We propose a novel framework that utilizes the robustness value of STL specifications to guide the trajectory modeling process in DT.
Our empirical results demonstrate that SDT is capable of learning a safe and high-reward policy in challenging offline safe RL tasks that involve temporal and logical requirements, and can adapt to different target suffixes without re-training and performs effectively across diverse target suffix configurations.
%These advantages make SDT preferable for real-world applications with complex safety constraints.
Future works will explore the use of STL to specify both safety and performance objectives in DT.
% \li{a sentence on future work if we have the space}

% There are also several limitations of SDT: 1) it requires more computing resources due to the Transformers architecture; 2) it lacks rigorous theoretical guarantees for specification satisfaction; 3) improper target reward and suffix can still deteriorate the performance; 
% Nevertheless, we hope our findings can provide fresh insights for extending the application of RL via supervised learning to broader domains.

% \vspace{-1mm}
\section*{Acknowledgement}
The authors thank the anonymous reviewers for their invaluable feedback and constructive suggestions.
This material is based upon work supported by the National Science Foundation under Grant No. CCF-2340776.

\vspace{-1mm}
\section*{Impact Statement}
This paper presents a novel framework within the realm of reinforcement learning, aiming to advance the field through innovative approaches and applications. 
First of all, the methods, experiments, and results outlined in this paper do not pose any ethical concerns. 
Secondly, it's crucial for researchers to proceed with care, especially when setting specifications and conducting tests in real-world settings, since misspecified specifications may result in serious and unforeseen consequences.
Lastly, we hope our findings can provide fresh insights for extending the application of reinforcement learning to broader domains.

% \clearpage
\bibliography{icml2024}
\bibliographystyle{icml2024}

%%%%%%%%%%%%%%%%%%%%%%%%%%%%%%%%%%%%%%%%%%%%%%%%%%%%%%%%%%%%%%%%%%%%%%%%%%%%%%%
%%%%%%%%%%%%%%%%%%%%%%%%%%%%%%%%%%%%%%%%%%%%%%%%%%%%%%%%%%%%%%%%%%%%%%%%%%%%%%%
% APPENDIX
%%%%%%%%%%%%%%%%%%%%%%%%%%%%%%%%%%%%%%%%%%%%%%%%%%%%%%%%%%%%%%%%%%%%%%%%%%%%%%%
%%%%%%%%%%%%%%%%%%%%%%%%%%%%%%%%%%%%%%%%%%%%%%%%%%%%%%%%%%%%%%%%%%%%%%%%%%%%%%%
\newpage
\appendix
\onecolumn

% You can have as much text here as you want. The main body must be at most $8$ pages long.
% For the final version, one more page can be added.
% If you want, you can use an appendix like this one.  

% The $\mathtt{\backslash onecolumn}$ command above can be kept in place if you prefer a one-column appendix, or can be removed if you prefer a two-column appendix.  Apart from this possible change, the style (font size, spacing, margins, page numbering, etc.) should be kept the same as the main body.
% In the appendix, we present more details on the implementation, results, and analysis.

% \section{Environment Setting}
% \vspace{-3mm}
\section{Environment Setting}
\label{app:envs}
% \subsection{Experiment Description}

\textbf{Reward and cost functions defined in the environments.}
We use the \texttt{Bullet-safety-gym}~\cite {gronauer2022bullet} environments for this set of experiments. 
In the \texttt{Run} environments, agents receive rewards for high-speed movement between two safety boundaries. However, they incur penalties when they either cross these boundaries or surpass a velocity threshold that is specific to different types of robots.
The reward and cost function are defined as:
\begin{align*}
    % \small
    r(\bm{s_t}) & = \frac{-y_t v_x + x_t v_y}{1 + | |\sqrt{x_t^2+y_t^2}-r|} \\
    c(\bm{s_t}) & = \bm{1}(|x| > x_{\text{lim}})
\end{align*}
where $\bm{s}_t = [x_t, y_t, v_x, v_y]$, $r$ is the radius of the circle, and $x_{\text{lim}}$ specifies the range of the safety region.
In the \texttt{Circle} environments, agents gain rewards for circular motion in a clockwise direction but are required to remain inside a designated safe area, which is smaller than the circumference of the intended circle.
The reward and cost functions are defined as:
\begin{align*}
    r(\bm{s_t}) & = ||\bm{x_{t-1}}-\bm{g}||_2 - ||\bm{x}_t-\bm{g}||_2 \\
    c(\bm{s_t}) & = \bm{1} (|y| > y_{\text{lim}} \text{ or } ||\bm{v_t}||_2 > v_{\text{lim}})
\end{align*}
where $y_{\text{lim}}$ is the safety boundary and $v_{\text{lim}}$ is the velocity limit.
To establish the criteria that zero cost means no violations of the specification, we relabel the cost.
For the \texttt{Circle} environment, the relabeled cost of the current state is 1 if the costs of its previous 5 steps are all 1. 
For the \texttt{Run} environment, the relabeled cost of the current state is 1 if the costs related to speeding of its previous 5 steps are all 1 or the cost related to safe boundary crossing is 1.

\begin{figure}[ht]
\centering
\includegraphics[width=1.\linewidth]{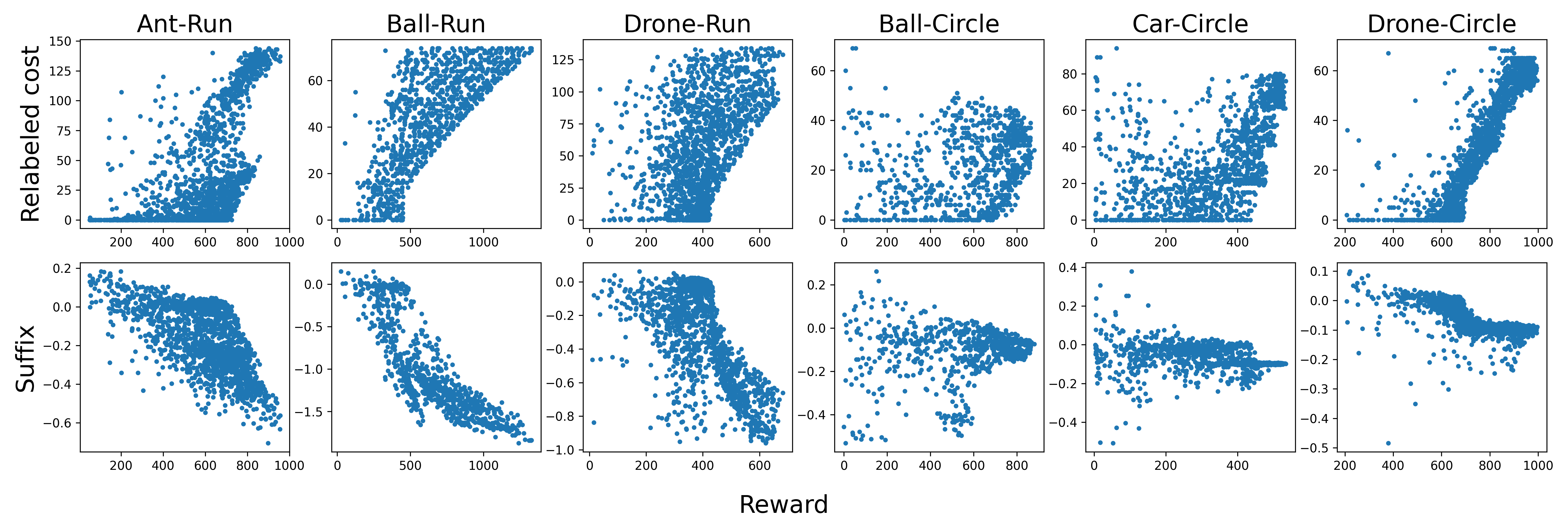}
\vspace{-3mm}
\caption{
Illustration of the offline dataset.
% The first-row plots show the relabeled cost versus reward, the second-row plots show the suffix versus reward, and the last-row plots show different target suffix configurations during evaluation in the ablation study.
The first-row plots show the relabeled cost versus reward and the second-row plots show the suffix ($\rho_{suf}(\tau_{1:T}, 1, \phi)$) versus reward.
Each column represents an environment.
Each point denotes a collected trajectory (not necessarily to be unique) with corresponding episodic relabeled cost (or suffix) and reward value.
% Each point denotes a trajectory.
}
\label{fig:data-vis}
\end{figure}

\label{app:data-vis}
\textbf{Offline dataset visualization.}
The dataset suffix-reward and relabeled cost-reward plots for the training tasks \texttt{Ant-Run}, \texttt{Ball-Run}, \texttt{Drone-Run}, \texttt{Ball-Circle}, \texttt{Car-Circle}, and \texttt{Drone-Circle}, are shown in Figure.~\ref{fig:data-vis}. 
Analyzing the figures provided, we can generally discern an increasing trend for the reward in relation to the cost. 
In other words, as cost increases, so too might the reward return,
underscoring the inherent trade-off between reward and cost. 
This phenomenon aligns with findings discussed in previous works~\cite{liu2023constrained, liu2022robustness}, which is also
one of the reasons that SDT is evaluated in \texttt{Bullet-Safety-Gym} environments.
In contrast, the same clear increasing trend is not observable in \texttt{SafetyGymnasium} environments~\cite{ray2019benchmarking, liu2023datasets, ji2023omnisafe}, such as \texttt{Goal}, \texttt{Button}, and \texttt{Push}. 
Moreover, highly stochastic environments pose unique challenges to the RCSL algorithms~\cite{paster2022you} and are beyond the scope of this paper.

\textbf{SDT and baselines implementation.}
Our implementation of SDT is built on the public codebase provided by the \texttt{OSRL} library\footnote{https://github.com/liuzuxin/OSRL}, which offers a collection of elegant and extensible implementations of state-of-the-art offline safe RL algorithms.
We use the \texttt{STLCG} toolbox\footnote{https://github.com/StanfordASL/stlcg} to compute the robustness value of the specification and add the corresponding prefix and suffix tokens as input to the transformers.
% For RvS and its variants, we implement them within the framework of the library and use the same 
In our experiments, we train SDT and baselines 200000 steps to ensure convergence and keep the rest of the hyperparameters used in the library unchanged. 
The constrained optimization RL baselines are trained and evaluated using a cost threshold of $d=0$.
During evaluation, the target cost is 0 for CDT and RvS-RC. 
The target suffix is $\{ 0.02, 0.01, 0.02, 0.09, 0.06, 0.02, 0.06, 0.06, 0.04 \}$ for SDT and RvS-R$\rho$ and the target reward is $\{ 720, 440, 410, 610, 400, 650, 300, 300, 600 \}$ for SDT and the conditioned RL baselines in \texttt{Ant-Run}, \texttt{Ball-Run}, \texttt{Drone-Run}, \texttt{Ball-Circle}, \texttt{Car-Circle}, \texttt{Drone-Circle}, \texttt{Ball-Reach}, \texttt{Car-Reach}, and \texttt{Drone-Reach},
respectively.
% \section{Experiment Setting and Hyperparameters}
% \label{app:envs}
% % \subsection{Experiment Description}

% We use the \texttt{Bullet-safety-gym}~\cite {gronauer2022bullet} environments for this set of experiments. 
% In the \texttt{Run} environments, agents receive rewards for high-speed movement between two safety boundaries. However, they incur penalties when they either cross these boundaries or surpass a velocity threshold that is specific to different types of robots.
% The reward and cost function are defined as:
% \begin{align*}
%     r(\bm{s_t}) & = \frac{-y_t v_x + x_t v_y}{1 + | |\sqrt{x_t^2+y_t^2}-r|} \\
%     c(\bm{s_t}) & = \bm{1}(|x| > x_{\text{lim}})
% \end{align*}
% where $\bm{s}_t = [x_t, y_t, v_x, v_y]$, $r$ is the radius of the circle, and $x_{\text{lim}}$ specifies the range of the safety region.
% In the \texttt{Circle} environments, agents gain rewards for circular motion in a clockwise direction but are required to remain inside a designated safe area, which is smaller than the circumference of the intended circle.
% The reward and cost functions are defined as:
% \begin{align*}
%     r(\bm{s_t}) & = ||\bm{x_{t-1}}-\bm{g}||_2 - ||\bm{x}_t-\bm{g}||_2 \\
%     c(\bm{s_t}) & = \bm{1} (|y| > y_{\text{lim}} \text{ or } ||\bm{v_t}||_2 > v_{\text{lim}})
% \end{align*}
% where $y_{\text{lim}}$ is the safety boundary and $v_{\text{lim}}$ is the velocity limit.

% \subsection{Hyperparameters}
% \label{app:params}

\section{Complete Results of SDT}
\label{app:exp-results}
The full evaluation results for different trained policies are presented in Table \ref{tab:full-results}. 
The columns of average performance are the same as Table~\ref{tab:temporal-exp-results}.
All values are averaged among 3 random seeds and 20 trajectories for each seed. 
SDT, CDT, RvS-R$\rho$, and RvS-RC are all evaluated using the same target reward. 
Both SDT and RvS-R$\rho$ undergo evaluation with the same target suffix, while CDT, RvS-RC, and other baseline methods are examined using the same target cost of zero.
% Due to the page limit, the complete results are listed in Appendix~\ref{app:exp-results}.
BC-Safe is fed with solely the zero-violation trajectories, but it fails to learn zero-violation policies and exhibits conservative performance and low reward. 
Our method shows high satisfaction rates, suggesting it consistently adheres to the safety specifications throughout various environments. 
RvS-R$\rho$ also performs well but cannot realize consistent performance in all environments, e.g. in \texttt{Ball-Run} environment, it only has marginal improvement over BC-safe.
One reason is that RvS-R$\rho$ does not use sequential modeling as transformers and thus falls short of learning temporal policies.
The Q-learning-based algorithms, including BCQ-Lag, BEAR-Lag, and CPQ, as well as COptiDICE, vacillate between excessive conservatism and riskiness. 
For example, CPQ obtains perfect satisfaction in \texttt{Car-Cricle} environment but achieves zero satisfaction in \texttt{Ball-Run} and \texttt{Drone-Run} environments; BEAR-Lag shows high rewards but also high costs in \texttt{Circle} environemnts, suggesting it takes more risks that could lead to unsafe outcomes.
However, they have the stitching ability and achieve higher rewards than the conditioned RL baselines whose reward is less than the maximum reward of safe trajectories.

\begin{table}[ht]
\centering
\renewcommand{\arraystretch}{1.4}
\resizebox{1.\textwidth}{!}{
\begin{tabular}{|c|ccc|ccc|ccc|ccc|}
\hline
                          & \multicolumn{3}{c|}{Ant-Run}                                                                                       & \multicolumn{3}{c|}{Ball-Run}                                                                                        & \multicolumn{3}{c|}{Drone-Run}                                                                                       & \multicolumn{3}{c|}{Average}                                                                                         \\
\multirow{-2}{*}{Methods} & Reward $\uparrow$                    & Cost $\downarrow$                    & Rate $\uparrow$                      & Reward $\uparrow$                    & Cost $\downarrow$                      & Rate $\uparrow$                      & Reward $\uparrow$                    & Cost $\downarrow$                      & Rate $\uparrow$                      & Reward $\uparrow$                    & Cost $\downarrow$                      & Rate $\uparrow$                      \\ \hline
SDT(ours)                 & \textbf{0.95$\pm$0.01}               & \textbf{0.0$\pm$0.0}                 & \textbf{1.0$\pm$0.0}                 & 0.97$\pm$0.01                        & \textbf{0.0$\pm$0.0}                   & \textbf{1.0$\pm$0.0}                 & {\color[HTML]{656565} 0.97$\pm$0.01} & {\color[HTML]{656565} 0.98$\pm$3.43}   & {\color[HTML]{656565} 0.92$\pm$0.02} & \textbf{0.96$\pm$0.01}               & 0.33$\pm$2.04                          & \textbf{0.97$\pm$0.04}               \\
RvS-R$\rho$               & 0.88$\pm$0.1                         & \textbf{0.0$\pm$0.0}                 & \textbf{1.0$\pm$0.0}                 & 0.56$\pm$0.37                        & 22.62$\pm$26.14                        & 0.5$\pm$0.41                         & {\color[HTML]{656565} 0.95$\pm$0.06} & {\color[HTML]{656565} 4.07$\pm$16.12}  & {\color[HTML]{656565} 0.87$\pm$0.02} & 0.79$\pm$0.28                        & 8.89$\pm$20.28                         & 0.79$\pm$0.32                        \\ \hline
CDT                       & \textbf{0.95$\pm$0.02}               & 0.07$\pm$0.4                         & 0.97$\pm$0.05                        & \textbf{0.98$\pm$0.01}               & 0.63$\pm$2.44                          & 0.85$\pm$0.18                        & {\color[HTML]{656565} 0.96$\pm$0.05} & {\color[HTML]{656565} 0.25$\pm$1.27}   & {\color[HTML]{656565} 0.95$\pm$0.04} & \textbf{0.96$\pm$0.03}               & \textbf{0.32$\pm$1.62}                 & 0.92$\pm$0.12                        \\
RvS-RC                    & {\color[HTML]{656565} 0.87$\pm$0.17} & {\color[HTML]{656565} 0.43$\pm$0.99} & {\color[HTML]{656565} 0.8$\pm$0.15}  & {\color[HTML]{656565} 2.02$\pm$0.56} & {\color[HTML]{656565} 49.9$\pm$19.22}  & {\color[HTML]{656565} 0.03$\pm$0.05} & {\color[HTML]{656565} 0.91$\pm$0.1}  & {\color[HTML]{656565} 10.68$\pm$20.09} & {\color[HTML]{656565} 0.75$\pm$0.32} & {\color[HTML]{656565} 1.27$\pm$0.63} & {\color[HTML]{656565} 20.34$\pm$26.69} & {\color[HTML]{656565} 0.53$\pm$0.41} \\
BC-safe                   & 0.92$\pm$0.06                        & 0.25$\pm$0.77                        & 0.88$\pm$0.05                        & 0.52$\pm$0.36                        & 9.18$\pm$12.72                         & 0.48$\pm$0.24                        & \textbf{0.88$\pm$0.12}               & \textbf{0.12$\pm$0.78}                 & \textbf{0.97$\pm$0.02}               & 0.78$\pm$0.29                        & 3.18$\pm$8.5                           & 0.78$\pm$0.25                        \\ \hline
BCQ-Lag                   & {\color[HTML]{656565} 0.8$\pm$0.18}  & {\color[HTML]{656565} 2.78$\pm$6.03} & {\color[HTML]{656565} 0.67$\pm$0.31} & {\color[HTML]{656565} 0.83$\pm$0.42} & {\color[HTML]{656565} 10.68$\pm$16.04} & {\color[HTML]{656565} 0.22$\pm$0.27} & {\color[HTML]{656565} 1.06$\pm$0.16} & {\color[HTML]{656565} 76.77$\pm$32.92} & {\color[HTML]{656565} 0.02$\pm$0.02} & {\color[HTML]{656565} 0.9$\pm$0.3}   & {\color[HTML]{656565} 30.08$\pm$39.49} & {\color[HTML]{656565} 0.3$\pm$0.36}  \\
BEAR-Lag                  & 0.01$\pm$0.03                        & \textbf{0.0$\pm$0.0}                 & \textbf{1.0$\pm$0.0}                 & {\color[HTML]{656565} 1.91$\pm$1.32} & {\color[HTML]{656565} 87.67$\pm$0.47}  & {\color[HTML]{656565} 0.0$\pm$0.0}   & {\color[HTML]{656565} 0.63$\pm$0.63} & {\color[HTML]{656565} 93.23$\pm$31.76} & {\color[HTML]{656565} 0.0$\pm$0.0}   & {\color[HTML]{656565} 0.85$\pm$1.16} & {\color[HTML]{656565} 60.3$\pm$46.47}  & {\color[HTML]{656565} 0.33$\pm$0.47} \\
CPQ                       & 0.03$\pm$0.05                        & \textbf{0.0$\pm$0.0}                 & \textbf{1.0$\pm$0.0}                 & {\color[HTML]{656565} 2.65$\pm$1.26} & {\color[HTML]{656565} 64.33$\pm$29.94} & {\color[HTML]{656565} 0.0$\pm$0.0}   & {\color[HTML]{656565} 0.81$\pm$0.38} & {\color[HTML]{656565} 86.53$\pm$45.67} & {\color[HTML]{656565} 0.0$\pm$0.0}   & {\color[HTML]{656565} 1.16$\pm$1.33} & {\color[HTML]{656565} 50.29$\pm$48.38} & {\color[HTML]{656565} 0.33$\pm$0.47} \\
COptiDICE                 & 0.78$\pm$0.07                        & 0.88$\pm$3.7                         & 0.9$\pm$0.04                         & {\color[HTML]{656565} 0.87$\pm$0.04} & {\color[HTML]{656565} 5.0$\pm$4.55}    & {\color[HTML]{656565} 0.33$\pm$0.47} & {\color[HTML]{656565} 1.05$\pm$0.08} & {\color[HTML]{656565} 59.62$\pm$10.71} & {\color[HTML]{656565} 0.0$\pm$0.0}   & {\color[HTML]{656565} 0.9$\pm$0.13}  & {\color[HTML]{656565} 21.83$\pm$27.68} & {\color[HTML]{656565} 0.41$\pm$0.46} \\ \hline

% \hhline{|=|===|===|===|===|}
\hhline{=============}
% \hline
                          & \multicolumn{3}{c|}{Ball-Circle}                                                                                    & \multicolumn{3}{c|}{Car-Circle}                                                                                      & \multicolumn{3}{c|}{Drone-Circle}                                                                                   & \multicolumn{3}{c|}{Average}                                                                                         \\
\multirow{-2}{*}{Methods} & Reward $\uparrow$                    & Cost $\downarrow$                     & Rate $\uparrow$                      & Reward $\uparrow$                    & Cost $\downarrow$                      & Rate $\uparrow$                      & Reward $\uparrow$                    & Cost $\downarrow$                     & Rate $\uparrow$                      & Reward $\uparrow$                    & Cost $\downarrow$                      & Rate $\uparrow$                      \\ \hline
SDT(ours)                 & 0.86$\pm$0.03                        & \textbf{0.48$\pm$1.02}                & \textbf{0.77$\pm$0.06}               & \textbf{0.86$\pm$0.02}               & 2.27$\pm$6.15                          & 0.85$\pm$0.07                        & 0.94$\pm$0.12                        & \textbf{0.07$\pm$0.4}                 & \textbf{0.97$\pm$0.02}               & 0.89$\pm$0.08                        & \textbf{0.94$\pm$3.73}                 & \textbf{0.86$\pm$0.1}                \\
RvS-R$\rho$               & 0.75$\pm$0.06                        & 0.87$\pm$1.59                         & 0.68$\pm$0.15                        & 0.76$\pm$0.09                        & 1.93$\pm$4.85                          & 0.82$\pm$0.09                        & 0.93$\pm$0.03                        & 0.47$\pm$1.8                          & 0.9$\pm$0.07                         & 0.81$\pm$0.11                        & 1.09$\pm$3.19                          & 0.8$\pm$0.14                         \\ \hline
CDT                       & \textbf{0.89$\pm$0.02}               & 1.08$\pm$1.42                         & 0.5$\pm$0.07                         & {\color[HTML]{656565} 0.91$\pm$0.01} & {\color[HTML]{656565} 4.53$\pm$6.37}   & {\color[HTML]{656565} 0.42$\pm$0.08} & \textbf{0.95$\pm$0.02}               & 1.07$\pm$1.52                         & 0.55$\pm$0.04                        & \textbf{0.92$\pm$0.03}               & 2.23$\pm$4.2                           & 0.49$\pm$0.09                        \\
RvS-RC                    & 0.72$\pm$0.21                        & 6.93$\pm$24.05                        & 0.6$\pm$0.15                         & 0.71$\pm$0.28                        & 13.42$\pm$28.14                        & 0.57$\pm$0.02                        & 0.83$\pm$0.27                        & 8.6$\pm$35.25                         & 0.68$\pm$0.05                        & 0.75$\pm$0.26                        & 9.65$\pm$29.64                         & 0.62$\pm$0.1                         \\
BC-safe                   & 0.59$\pm$0.23                        & 2.22$\pm$3.55                         & 0.5$\pm$0.15                         & 0.41$\pm$0.28                        & 12.45$\pm$15.38                        & 0.48$\pm$0.14                        & 0.86$\pm$0.13                        & 2.58$\pm$3.86                         & 0.5$\pm$0.11                         & 0.62$\pm$0.28                        & 5.75$\pm$10.51                         & 0.49$\pm$0.13                        \\ \hline
BCQ-Lag                   & {\color[HTML]{656565} 0.95$\pm$0.14} & {\color[HTML]{656565} 19.78$\pm$7.58} & {\color[HTML]{656565} 0.0$\pm$0.0}   & {\color[HTML]{656565} 0.66$\pm$0.27} & {\color[HTML]{656565} 21.8$\pm$17.81}  & {\color[HTML]{656565} 0.25$\pm$0.11} & {\color[HTML]{656565} 1.37$\pm$0.04} & {\color[HTML]{656565} 58.38$\pm$5.96} & {\color[HTML]{656565} 0.0$\pm$0.0}   & {\color[HTML]{656565} 0.99$\pm$0.34} & {\color[HTML]{656565} 33.32$\pm$21.25} & {\color[HTML]{656565} 0.08$\pm$0.13} \\
BEAR-Lag                  & {\color[HTML]{656565} 1.01$\pm$0.12} & {\color[HTML]{656565} 15.35$\pm$8.45} & {\color[HTML]{656565} 0.0$\pm$0.0}   & {\color[HTML]{656565} 0.94$\pm$0.15} & {\color[HTML]{656565} 38.8$\pm$16.71}  & {\color[HTML]{656565} 0.0$\pm$0.0}   & {\color[HTML]{656565} 1.27$\pm$0.08} & {\color[HTML]{656565} 45.45$\pm$12.4} & {\color[HTML]{656565} 0.0$\pm$0.0}   & {\color[HTML]{656565} 1.07$\pm$0.18} & {\color[HTML]{656565} 33.2$\pm$18.3}   & {\color[HTML]{656565} 0.0$\pm$0.0}   \\
CPQ                       & 0.79$\pm$0.05                        & 2.28$\pm$4.01                         & 0.67$\pm$0.47                        & \textbf{0.86$\pm$0.03}               & \textbf{0.0$\pm$0.0}                   & \textbf{1.0$\pm$0.0}                 & 0.12$\pm$0.17                        & 14.92$\pm$30.51                       & 0.5$\pm$0.22                         & 0.59$\pm$0.35                        & 5.73$\pm$18.94                         & 0.72$\pm$0.36                        \\
COptiDICE                 & {\color[HTML]{656565} 0.82$\pm$0.09} & {\color[HTML]{656565} 8.28$\pm$2.93}  & {\color[HTML]{656565} 0.02$\pm$0.02} & {\color[HTML]{656565} 0.57$\pm$0.07} & {\color[HTML]{656565} 22.95$\pm$26.69} & {\color[HTML]{656565} 0.27$\pm$0.12} & {\color[HTML]{656565} 0.59$\pm$0.05} & {\color[HTML]{656565} 5.98$\pm$5.93}  & {\color[HTML]{656565} 0.22$\pm$0.06} & {\color[HTML]{656565} 0.66$\pm$0.13} & {\color[HTML]{656565} 12.41$\pm$17.57} & {\color[HTML]{656565} 0.17$\pm$0.13} \\ \hline
\hhline{=============}
\multirow{2}{*}{Methods} & \multicolumn{3}{c|}{Ball-Reach}                                               & \multicolumn{3}{c|}{Car-Reach}                                                 & \multicolumn{3}{c|}{Drone-Reach}                                               & \multicolumn{3}{c|}{Average}                                                  \\
                         & Reward $\uparrow$        & Cost $\downarrow$        & Rate $\uparrow$         & Reward $\uparrow$        & Cost $\downarrow$         & Rate $\uparrow$         & Reward $\uparrow$        & Cost $\downarrow$        & Rate $\uparrow$          & Reward $\uparrow$        & Cost $\downarrow$       & Rate $\uparrow$          \\ \hline
SDT(ours)                & \textbf{0.61 $\pm$ 0.03} & \textbf{0.01 $\pm$ 0.04} & 0.68 $\pm$ 0.02         & \textbf{0.72 $\pm$ 0.02} & \textbf{-0.01 $\pm$ 0.04} & 0.62 $\pm$ 0.18         & \textbf{0.91 $\pm$ 0.02} & \textbf{0.01 $\pm$ 0.01} & \textbf{0.98 $\pm$ 0.03} & \textbf{0.75 $\pm$ 0.13} & \textbf{0.0 $\pm$ 0.03} & \textbf{0.76 $\pm$ 0.19} \\ \hline
RvS-R$\rho$              & 0.55 $\pm$ 0.07          & \textbf{0.01 $\pm$ 0.03} & \textbf{0.7 $\pm$ 0.05} & 0.3 $\pm$ 0.18           & \textbf{-0.01 $\pm$ 0.12} & \textbf{0.75 $\pm$ 0.2} & 0.81 $\pm$ 0.14          & -0.02 $\pm$ 0.12         & 0.5 $\pm$ 0.25           & 0.55 $\pm$ 0.25          & -0.01 $\pm$ 0.1         & 0.65 $\pm$ 0.22          \\
BC-safe                  & 0.63 $\pm$ 0.26          & -0.06 $\pm$ 0.09         & 0.22 $\pm$ 0.02         & 0.41 $\pm$ 0.29          & -0.05 $\pm$ 0.19          & 0.45 $\pm$ 0.2          & 0.69 $\pm$ 0.29          & -0.01 $\pm$ 0.04         & 0.35 $\pm$ 0.05          & 0.58 $\pm$ 0.3           & -0.04 $\pm$ 0.12        & 0.34 $\pm$ 0.15          \\ \hline
\end{tabular}
}
\caption{
Complete evaluation results of the normalized reward, cost, and satisfaction rate. 
$\uparrow$: the higher the reward, the better.
$\downarrow$: the lower the cost (closer to 0), the better.
Agents with a higher satisfaction rate than BC-safe are considered safe. 
{\color[HTML]{656565} Gray}: unsafe agents.
\textbf{Bold}: the best in the respective metric among safe agents.
}
\label{tab:full-results}
\end{table}

\begin{figure}[ht]
\centering
\includegraphics[width=1.0\linewidth]{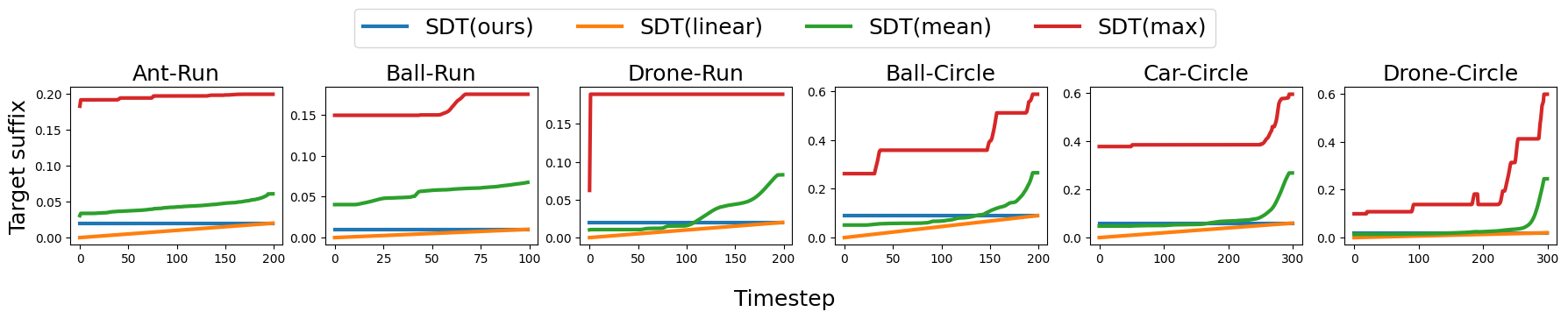}
\vspace{-6mm}
\caption{
Illustration of different target suffix configurations.
i) SDT(ours): fixed target suffix
ii) SDT(linear): linearly increasing target suffix; iii)  SDT(mean): average suffix from safe trajectories as target suffix; iiii) SDT(max): maximum suffix from safe trajectories as target suffix.
}
\label{fig:suffix-vis}
\end{figure}

\textbf{Target suffix configurations.}
The different target suffix configurations used in section~\ref{sec:ablation} are shown in Figure~\ref{fig:suffix-vis}.
Due to the $\mathbf{G}$ in the specification defined in Eq.~(\ref{eq:spec-run}) and (\ref{eq:spec-circle}), the suffix is monotonically increasing.
Recall that positive robustness values indicate that the specification is satisfied and a higher value indicates stronger satisfaction.
Therefore, we set positive robustness values as our target suffix for evaluation.
As shown in Figure~\ref{fig:data-vis}, the suffix associated with safe trajectories yielding high rewards is nearly 0, which aligns with the \textit{tempting} concept in~\cite{liu2022robustness}.
This suggests that aiming for both a high reward and a high target suffix may lead to reduced performance, as achieving such a combination is impractical.
In the environments, a high reward means navigation along the safety boundary, while a high suffix denotes maintaining distance from it.
This disparity explains why satisfaction rates of SDT(max) and SDT(mean) decrease compared to SDT(ours).
In the case of SDT(linear), a target suffix smaller than that in SDT(ours) is overwhelmed by a high target reward, thus resulting in violations of satisfaction.
Empirically, a fixed target suffix works best among these tested configurations.

\begin{figure*}[ht]
\centering
\includegraphics[width=0.9\linewidth]{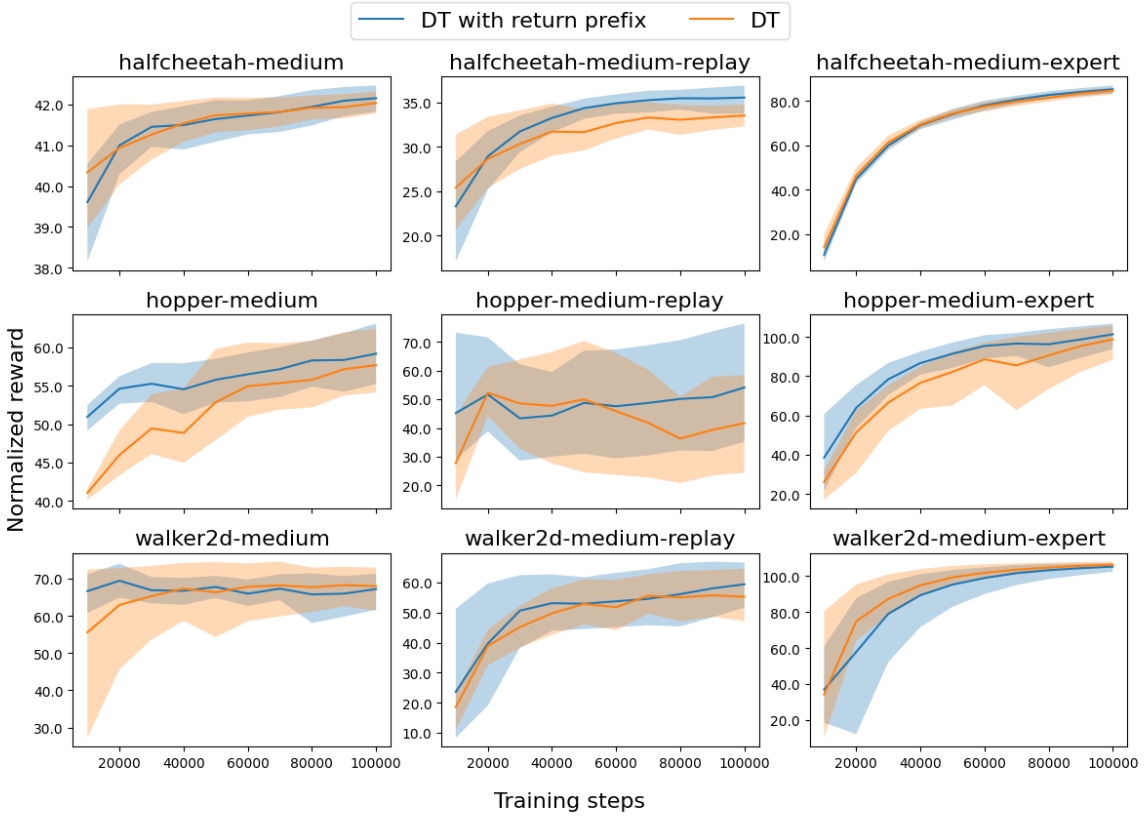}
\vspace{-2mm}
\caption{Evaluation results of DT and DT with reward prefix during training. 
The solid line and the light shade area represent the mean and mean $\pm$ standard deviation.
The normalized reward follows the evaluation protocol in \texttt{D4RL}.
}
\label{fig:app-dt-train}
\end{figure*}

\section{Experiments of Decision Transformers}
\label{app:dt-d4rl}
To demonstrate the role of the prefix, we evaluate the performance of the Decision Transformers (DT) and DT with reward prefix on the \texttt{D4RL} benchmark~\cite{fu2020d4rl}. 
We use the official codebase of DT\footnote{https://github.com/kzl/decision-transformer} and use the default hyperparameters to train \texttt{Halfcheetah}, \texttt{Walker2D}, and \texttt{Hopper} on different types of dataset: \texttt{medium} that collected from a partially-trained policy; \texttt{medium-replay} that consists of recording all samples in the replay buffer observed during training until the policy reaches the medium level of performance; and \texttt{medium-expert} that contains equal amounts of expert demonstrations and suboptimal data.
DT and DT with reward prefix are evaluated every 10000 steps during training and the results are shown in Figure~\ref{fig:app-dt-train}.
The reward prefix is beneficial not only to the performance after convergence but also to facilitate the training process for most of the tasks since DT with reward prefix achieves higher reward.
Another observation is that the impact of the reward prefix is obvious in \texttt{medium} and \texttt{medium-replay} datasets, while the improvement is marginal in \texttt{medium-expert} datasets, indicating that the reward prefix has greater importance in learning good policies from suboptimal data.
Empirically, the better performance of introducing the reward prefix supports our augments in section~\ref{sec:stl-rob} that the prefix is not redundant to the suffix but provides additional information that promotes policy learning.

% $\mathbf{P}_{pre} = \{ \rho_{pre}(\tau, t-K, \phi)$, $...$, $\rho_{pre}(\tau, t, \phi) \}$, suffix $\mathbf{P}_{suf} = \{ \rho_{suf}(\tau, t-K, \phi)$, $...$, $\rho_{suf}(\tau, t, \phi) \}$, 
%%%%%%%%%%%%%%%%%%%%%%%%%%%%%%%%%%%%%%%%%%%%%%%%%%%%%%%%%%%%%%%%%%%%%%%%%%%%%%%
%%%%%%%%%%%%%%%%%%%%%%%%%%%%%%%%%%%%%%%%%%%%%%%%%%%%%%%%%%%%%%%%%%%%%%%%%%%%%%%

\end{document}